\documentclass{article}
\usepackage{log_2024}						

\usepackage{booktabs}						
\usepackage{multirow}						
\usepackage{amsfonts}						
\usepackage{graphicx}						
\usepackage{duckuments}						

\usepackage[numbers,compress,sort]{natbib}	

\usepackage{microtype}
\usepackage{graphicx}
\usepackage{subfigure}
\usepackage{booktabs} 
\usepackage{wrapfig}
\usepackage{enumitem}
\usepackage[utf8]{inputenc} 
\usepackage[T1]{fontenc}    
\usepackage{url}            
\usepackage{booktabs}       
\usepackage{amsfonts}       
\usepackage{nicefrac}       
\usepackage{microtype}      
\usepackage{xcolor}         
\usepackage{amsmath}
\usepackage[numbers,compress,sort]{natbib}	

\usepackage[textsize=tiny]{todonotes}
\newcommand{\ppa}{{\small\textsc{ogbg-ppa}}}
\newcommand{\molhiv}{{\small\textsc{ogbg-molhiv}}}
\newcommand{\collab}{{\small\textsc{ogbl-collab}}}
\newcommand{\molpcba}{{\small\textsc{ogbg-molpcba}}}
\newcommand{\PCQ}{{\small\textsc{pcqm4mv2}}}
\newcommand{\reddit}{{\small\textsc{reddit-multi-12k}}}
\newcommand{\arxiv}{{\small\textsc{ogbn-arxiv}}}
\newcommand{\products}{{\small\textsc{ogbn-products}}}
\newcommand{\proteins}{{\small\textsc{ogbn-proteins}}}
\newcommand{\linkppa}{{\small\textsc{ogbl-ppa}}}
\newcommand{\cit}{{\small\textsc{ogbl-citation2}}}

\title{Towards Neural Scaling Laws on Graphs}

\author[J. Liu et al.]{%
Jingzhe Liu\\
Michigan State University\\
\email{liujin33@msu.edu}\And
Haitao Mao\\
Michigan State University\\
\email{haitaoma@msu.edu}\And
Zhikai Chen\\
Michigan State University\\
\email{chenzh85@msu.edu}\And
Tong Zhao\\
Snap Inc\\
\email{tzhao@snap.com}\And
Neil Shah\\
Snap Inc\\
\email{nshah}@snap.com\And
Jiliang Tang\\
Michigan State University\\
\email{tangjili@msu.edu}
}

\begin{document}

\maketitle

\begin{abstract}

Deep graph models (e.g., graph neural networks and graph transformers) have become important techniques for leveraging knowledge across various types of graphs. 
Yet, the neural scaling laws on graphs, i.e., how the performance of deep graph models changes with model and dataset sizes, have not been systematically investigated, casting doubts on the feasibility of achieving large graph models. 
To fill this gap, we benchmark many graph datasets from different tasks and make an attempt to establish the neural scaling laws on graphs from both model and data perspectives. 
The model size we investigated is up to 100 million parameters, and the dataset size investigated is up to 50 million samples.
We first verify the validity of such laws on graphs, establishing proper formulations to describe the scaling behaviors. 
For model scaling, we identify that despite the parameter numbers, the model depth also plays an important role in affecting the model scaling behaviors, which differs from observations in other domains such as CV and NLP. 
For data scaling, we suggest that the number of graphs can not effectively measure the graph data volume in scaling law since the sizes of different graphs are highly irregular. Instead, we reform the data scaling law with the number of nodes or edges as the metric to address the irregular graph sizes.  
We further demonstrate that the reformed law offers a unified view of the data scaling behaviors for various fundamental graph tasks including node classification, link prediction, and graph classification. 
This work provides valuable insights into neural scaling laws on graphs, which can serve as an important tool for collecting new graph data and developing large graph models.
\end{abstract}

\section{Introduction\label{sec:intro}}

The neural scaling law~\cite{rosenfeld2019constructive,kaplan2020scaling,abnar2021exploring} serves as an important tool in the development of large models. It quantitatively describes how model performance grows with the scale of training, which have two basic forms: data scaling law~(how the performance changes with training dataset size) and model scaling law~(how the performance changes with model size).
Neural scaling laws guide the construction of large models from the following perspectives:
(1) They enable us to predict model performance on large-scale training by extrapolating from small-scale runs with lower computational budgets. For instance, the performance of GPT-4 on coding problems fits accurately with the prediction of the scaling law~\cite{achiam2023gpt}.
(2) They enable us to identify better model architectures through small-scale training, typically selecting those with the largest performance gains for a given computational overhead. 
(3) They serves as the guidance for resource allocation, i.e., they could tell us how to distribute the cost of data collection and computation for the given model to reach the expected performance.
Such laws have played an instrumental role in the recent successes of large models in natural language processing~(NLP) and computer vision~(CV) domains~\cite{brown2020language,touvron2023llama,chowdhery2023palm, Zhai_2022_CVPR, dehghani2023scaling}. 

\begin{figure}[ht!]
    \centering
    \subfigure{
    \begin{minipage}[b]{0.45\textwidth}
    \includegraphics[width=0.9\columnwidth]{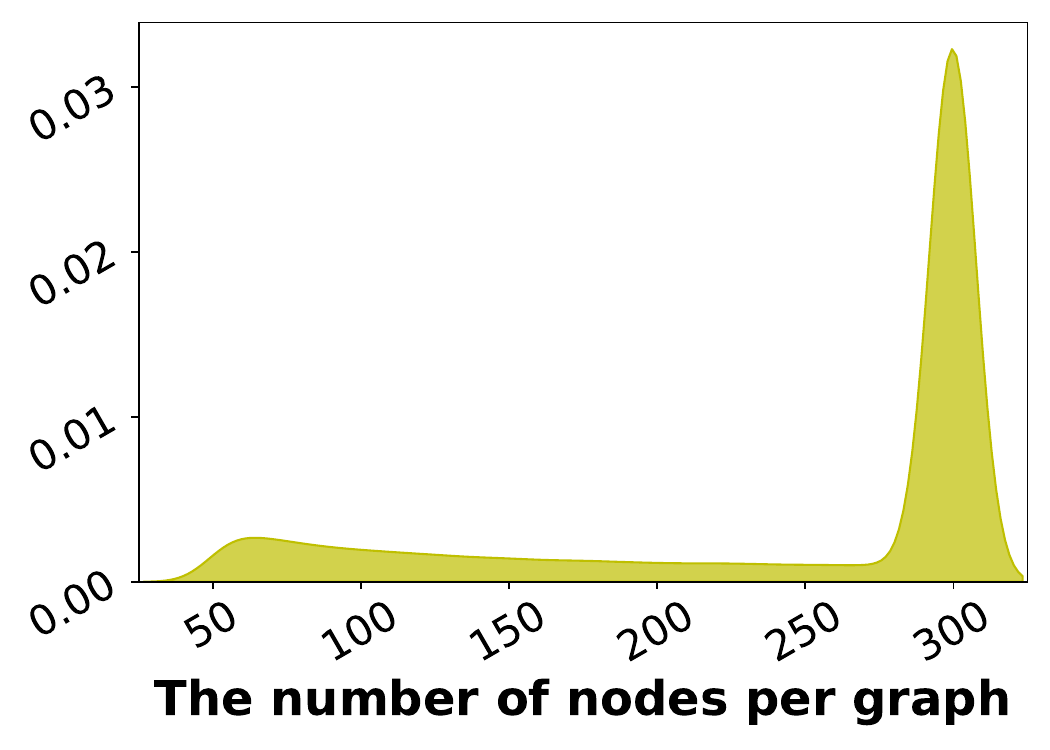}
    \end{minipage}
    }
    \subfigure{
    \begin{minipage}[b]{0.45\textwidth}
    \includegraphics[width=0.9\columnwidth]{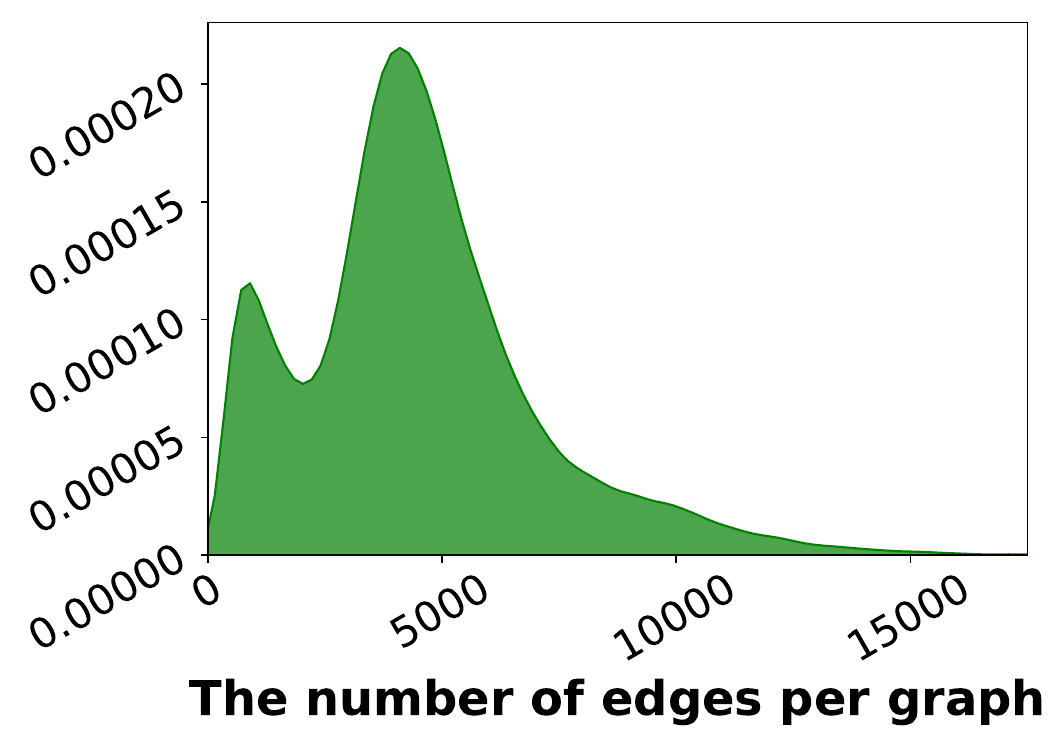}
    \end{minipage}
    }
    \vskip -1em
    \caption{The density distribution of the number of nodes and edges in \ppa{} dataset.}\label{fig:ppa info}
    \vskip -1.3em
\end{figure}

Despite the success of foundation models in NLP and CV, there is no existing effort investigating the neural scaling law in the graph machine learning domain. 
As an initial exploration, we found that such an investigation would meet unique challenges given the irregularity and diversity of graphs.
Specifically, both the number of nodes and edges of a graph can vary dramatically, different from the fixed-structure figures or texts in CV and NLP.
As an example shown in Figure \ref{fig:ppa info}, in the \ppa{} 
dataset of $158k$ graphs~\cite{szklarczyk2019string}, the number of nodes in each graph ranges from $25$ to $325$, and the number of edges from $54$ to $18,096$, indicating significant variability across the dataset. 
The variability in size suggests that the current metric (the number of graph samples) insufficiently represents the data volume, hinting at the need for more proper data metrics of the graph scaling laws. 
Moreover, the irregularity of graphs requires the deep graph models to introduce the message aggregation mechanisms. Hence, besides the model weights considered in the existing scaling law, deep graph models also need to consider the influence of the non-parametric aggregations, i.e., whether graph models with different numbers of aggregation layers could exhibit distinct scaling behaviors. 

In this work, we make a first attempt to address these unique challenges and develop the neural scaling laws on graphs to become a tool for building large graph models. 
Our investigation comes across a large range of domains and tasks in graph representation learning.
Experimental observations reveal neural scaling laws across different models and dataset sizes for graph learning tasks.

Our contributions can be summarized as follows:

\begin{itemize}[left=1em]
    \item We start our investigation with verifying the general forms of graph scaling laws. Specifically, we develop math formulas for the relation between model/dataset size and the model performance. Then we verify their validity with empirical studies across different datasets and models.
    \item We find that deep graph models with different numbers of layers will exhibit different model scaling behaviors, contrasting with the conclusions in CV and NLP domains. We then look deeper into the phenomenon and propose that the non-parametric aggregation operation could be one of the possible reasons.
    \item Regarding the size variability problem, we reveal that the number of nodes or edges is a better metric than the number of graphs when applying the data scaling law in graph classification tasks. To support this argument, we demonstrate that models trained on datasets with identical graph counts but different total numbers of edges exhibit distinctive scaling behaviors.
    \item We further extend the data scaling law to node classification and link prediction tasks, demonstrating the generality of using the number of nodes or edges as the data metric.
\end{itemize}





\textbf{Orgnizations.} 
The main focus of this work is on the two basic forms of neural scaling laws for the graph domain, i.e., the modeling scaling law and the data scaling law.
We tend to address the challenges caused by the irregularity of graph data when establishing the scaling laws.
Our investigations focus on the well-studied supervised graph learning while leaving the under-studied unsupervised learning tasks~\cite{sun2022does} as one future work. 
The following sections will be arranged as follows: 
In Section~\ref{sec:Background}, we briefly introduce the neural scaling laws and the related works.
In Section~\ref{sec:Setup}, we introduce the basic settings of our experiments.
In Section~\ref{sec:Basic Laws}, we verify the validity of the basic forms of neural scaling laws on the graphs. 
In Section~\ref{sec:The Effect of the number of model layers}, we reveal that model depth can affect the model scaling behaviors, which is a unique phenomenon in graph domains. 
In Section~\ref{sec:DataScaling}, we propose the number of edges is a suitable data metric when applying the data scaling law on graphs. 
To support the argument, we exhibit that models would have distinctions in scaling behavior when trained on datasets with different numbers of edges. 
Furthermore, we present that using the number of edges as the data metric could provide a unified data scaling law for various graph learning tasks, e.g., node classification, link prediction, and graph property prediction. 

\section{Related Work\label{sec:Background}}
 
\textbf{Functional forms of neural scaling laws.} Neural scaling laws give the quantitative relationship between the model performance and three main factors: the number of model parameters $\mathbf{N}$, the size of the training dataset $\mathbf{D}$, and the amount of computation $\mathbf{C}$. 
The relationship is generally described as a power-law, which has widely been applied in the domains of image classification~\cite{hestness2017deep, henighan2020scaling, sharma2022scaling, Zhai_2022_CVPR}, generative language modeling~\cite{kaplan2020scaling,henighan2020scaling,hoffmann2022training}, and natural language translation~\cite{gordon2021data,ghorbani2021scaling,pmlr-v162-bansal22b,fernandes2023scaling}.

Using the test error $\mathbf{\epsilon}$ as a metric for model performance, Hestness et al~\cite{hestness2017deep} propose the basic functional form of neural scaling laws as follows: 
\begin{equation}
    \mathbf{\epsilon  = aX^{-b} + \epsilon_\infty} \label{equation:error}
\end{equation}
The variable $\mathbf{X}$ represents any of the parameters $\mathbf{N}$, $\mathbf{D}$, or $\mathbf{C}$. The positive constants $\mathbf{a}$ and $\mathbf{b}$ govern the rate of error reduction as a function of $\mathbf{X}$, while $\mathbf{\epsilon_\infty} > 0$ specifies the minimum achievable error.
Equation~\ref{equation:error} is appropriate for regression tasks but is ill-defined for classification tasks, since the error rate $\mathbf{\epsilon_r}$ remains finite for models with zero training.
To fix this problem, Zhai et al~\cite{Zhai_2022_CVPR} proposes a modified function:
\begin{equation}
    \mathbf{\epsilon_r  = a(X+c)^{-b} + \epsilon_\infty} \label{equation:testerror}
\end{equation}
where a positive shift coefficient $\mathbf{c}$ is added to avoid having infinite $\mathbf{\epsilon_r}$ when $\mathbf{X}$ equals to zero. In this work, we will base our following investigations of graph classification tasks on this equation.

\textbf{The relationship between $\mathbf{N}$, $\mathbf{D}$ and overfitting.} The negative exponent $-\mathbf{b}$ in Equation~\ref{equation:testerror} indicates that the performance gain will be diminishing if only one of the parameters $\mathbf{N}$ and $\mathbf{D}$ is increased. To break this bottleneck, the dataset size and the model size need to be enlarged together. Specifically, Kaplan et al~\cite{kaplan2020scaling}
find that the model size and dataset size should increase together as $\mathbf{D\sim N^\alpha}$ ($0<\mathbf{\alpha}<1$) to attain optimal return for compute budgets. It is also observed that overfitting will likely occur if $\mathbf{N}$ surpasses $\mathbf{D^{1/\alpha}}$ significantly, which leads to negative performance gain.

\textbf{Graph learning and scaling.}
The graph is capable of representing non-Euclidean data with irregular and flexible structures, e.g., social networks~\cite{yanardag2015deep}, molecules~\cite{gilmer2017neural}, and protein networks~\cite{szklarczyk2019string}. 
Despite the popularity of graph machine learning, there are seldom large graph models. Notably, most deep GNNs remain small parameter size but more aggregation layers. Moreover, the neural scaling laws on graphs remain underexplored.
Some works have looked into the neural scaling laws for molecule representation learning~\cite{chen2023uncovering, sypetkowski2024scalability}, but the exploration is limited to the data and model types. It is unclear how the conclusion can be generalized to other graphs. To the best of our knowledge, we are the first to investigate neural scaling laws for the general graph domain.


\section{Preliminaries and Experiment Setups\label{sec:Setup}}




To ensure the applicability of neural scaling laws in the graph domain, our experiments are comprehensive to cover different types of models and graphs from various domains, e.g., molecule, biology, and academic. 
We investigate two major types of deep graph learning models, i.e., graph neural networks~(GNNs) and graph transformers. We explore 
GCN~\cite{kipf2016semi} and GIN~\cite{xu2018powerful} for GNNs, and SAT~\cite{pmlr-v162-chen22r}, and GraphGPS~\cite{rampavsek2022recipe} for graph transformers. Our datasets~\cite{morris2020tudataset,hu2020open,hu2021ogb} cover the the most common graph domains including social networks~(\reddit{}, \collab{}), molecules~(\molhiv{},\molpcba{},\PCQ{}), protein networks~(\ppa{}), and citation networks~(\arxiv{}).
Also, the chosen datasets are relatively large-scale ones among the academic datasets, ensuring valid observation of scaling behaviors.
Moreover, our experiments are conducted on three common graph tasks, i.e. graph property prediction, node classification, and link prediction, though graph property prediction is the focus.
Notably, accuracy rather than error rate is adopted in graph classification tasks as the default metric. To ensure a consistent observation, we switch the error rate $\epsilon_r$ in Equation~\ref{equation:testerror} to accuracy via applying the relation \textit{$\text{Accuracy} = 1 - \text{ErrorRate}$} and obtain the following equation:
\begin{equation}
    \mathbf{s  = -a(X+c)^{-b} + s_\infty} \label{equation:testscore}
\end{equation}
where $\mathbf{s}$ is the test score corresponding to specific metrics and $\mathbf{s_\infty = 1-\epsilon_\infty}$. Equation~\ref{equation:testscore} will be used in the following sections to fit the scaling law on classification tasks. The details of datasets and experimental settings can be found in Appendix~\ref{app:experiment settings}.

\section{Validity of Basic Scaling Laws on Graphs\label{sec:Basic Laws}}

We start our analysis by examining the basic forms of neural scaling laws in the graph domain. In other words, we aim to explore whether the model performance has a power-law relationship with the dataset size and model size. 
Hence, we first evaluate the performance of the models across various sizes and training set sizes.
Since the impact model depth is unclear yet, we vary the model size by

increasing the model width while fixing the model depth, and the effect of model depth will be further explored in Section~\ref{sec:The Effect of the number of model layers}.
To obtain subsets with different sizes, we randomly subsample the original training set by different sampling ratios.
Then, we fit the scaling law Equation~\ref{equation:error} and \ref{equation:testscore} to the empirical results on different model scales using the least square method.
To measure the fitting quality, the coefficient of determination $R^2$ is adopted as a metric. 

In this subsection, we focus on GIN and leave the results with similar observations on graph transformer in Section~\ref{sec:The Effect of the number of model layers} on GCN in Appendix~\ref{app:validity}.
Our experiments are conducted on two large graph property prediction datasets: \PCQ{} and \ppa{}. 

\begin{figure}[ht!]
    \centering
    \subfigure[Model scaling on \PCQ{}, $R^2=0.97$.\label{subfig:pcq model}]{
    
    \begin{minipage}[b]{0.45\textwidth}
    \includegraphics[width=\columnwidth]{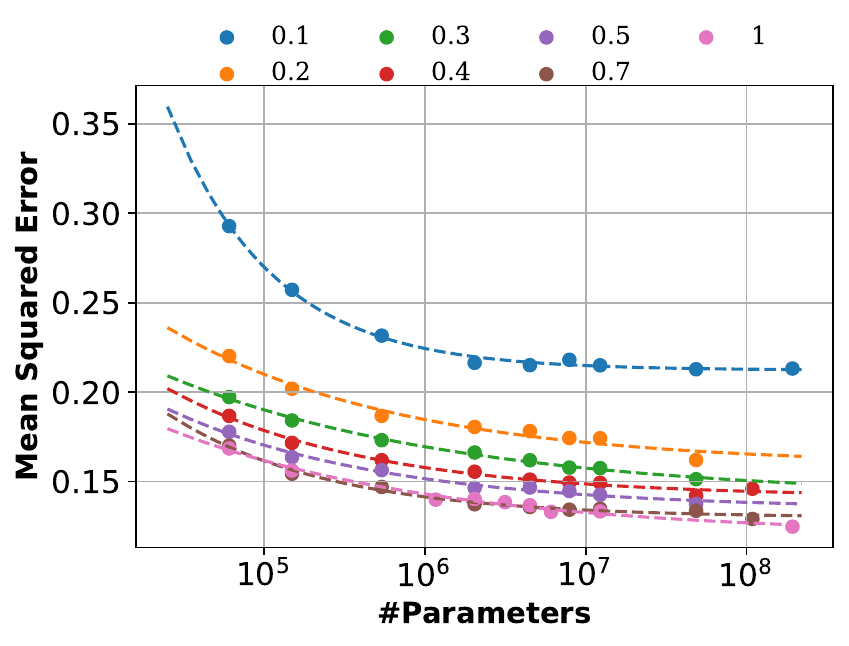}
    \end{minipage}
    }
    \subfigure[Data scaling on \PCQ{}, $R^2=0.98$.\label{subfig:pcq data}]{
    \begin{minipage}[b]{0.45\textwidth}
    \includegraphics[width=\columnwidth]{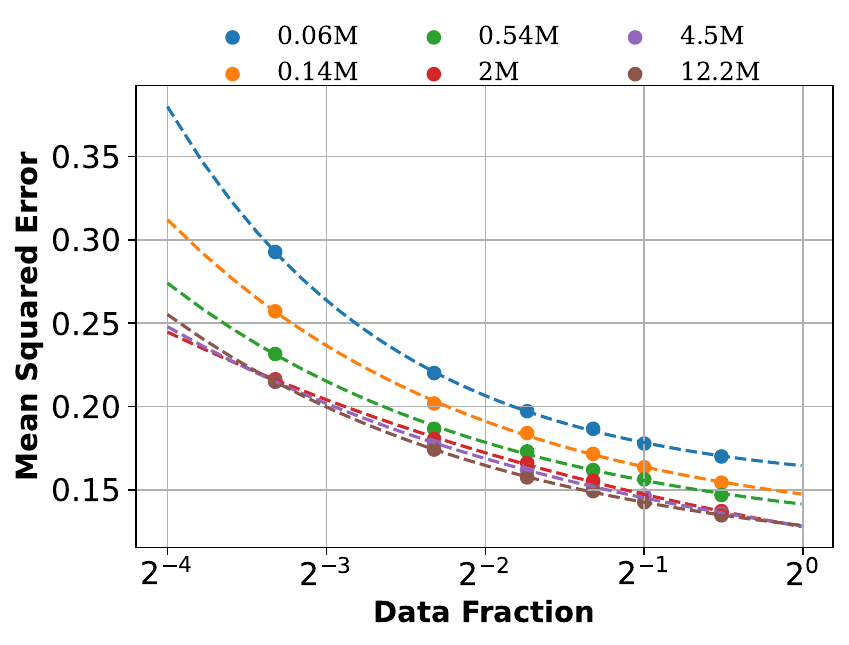}
    \end{minipage}
    }
    \vfill
 \subfigure[Model scaling on \ppa{}, $R^2=0.98$.\label{subfig:ppa model}]{

    \begin{minipage}[b]{0.45\textwidth}
    \includegraphics[width=\textwidth]{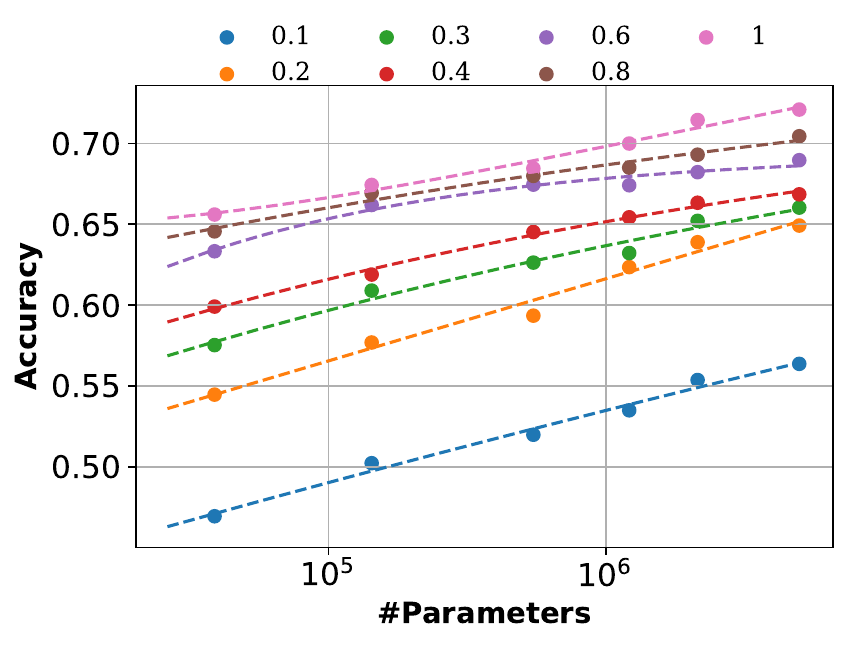}
    \end{minipage}
    }
    \subfigure[Data scaling on \ppa{}, $R^2=0.98$.\label{subfig:ppa data}]{

    \begin{minipage}[b]{0.45\textwidth}
    \includegraphics[width=\textwidth]{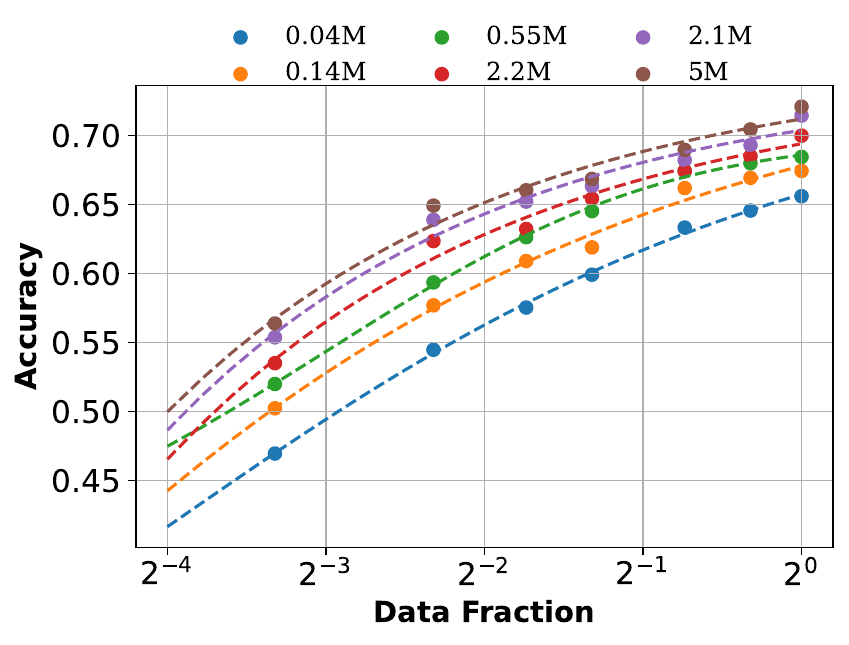}
    \end{minipage}
    }

    \caption{The model and data scaling behaviors of GIN on both \PCQ{} and \ppa{}. For model scaling, the color corresponds to the training data fraction; for data scaling, the color corresponds to the number of model parameters. The lower the mean square error is, the better the model performance is. The high $R^2$ values indicate that the scaling law curves fit well with the empirical values.}\label{fig:pcq scaling}

\end{figure}


We conduct experiments for both model scaling and data scaling behaviors on the graph regression dataset \PCQ{}. For regression tasks, the performance is measured by mean square error~(MSE), hence we use Equation~\ref{equation:error} to fit the empirical MSEs.
For model scaling, we fix the number of layers for the model to $6$ and vary the model width from $64$ to $1,024$. The experiments are repeated for different data fractions chosen from $\left[0.1,0.2,0.3,0.4,0.5,0.7,1\right]$.
The results, shown in Figure~\ref{subfig:pcq model}, depict each model's MSE as dots, with dashed lines representing fitted scaling law curves and colors indicating specific training data fractions.
We make the following observations from Figure~\ref{subfig:pcq model}: 
(1) The MSE tends to decrease when increasing the model scale and fixing the dataset scale.
(2) The MSE gradually decreases and approaches a lower bound $\epsilon$ asymptotically as the model becomes larger. 
The scaling curves corresponding to larger dataset sizes would have smaller $\epsilon$, i.e., higher upper bound of the model performance.
The observations in Figure~\ref{subfig:pcq data} exhibit similar patterns when scaling up data size while fixing the model size.
Generally speaking, both model scaling curves in Figure~\ref{subfig:pcq model} and data scaling curves in Figure~\ref{subfig:pcq data} fit well to empirical values with high $R^2$s, indicating that Equation~\ref{equation:error} could be leveraged to describe the neural scaling laws for graph regression tasks. 

\begin{figure}[ht!]
    \centering
    \subfigure[Scaling behaviors on \PCQ{}, $R^2=0.98$.\label{subfig:pcq combined}]{
    \begin{minipage}[b]{0.45\textwidth}
    \includegraphics[width=0.9\columnwidth]{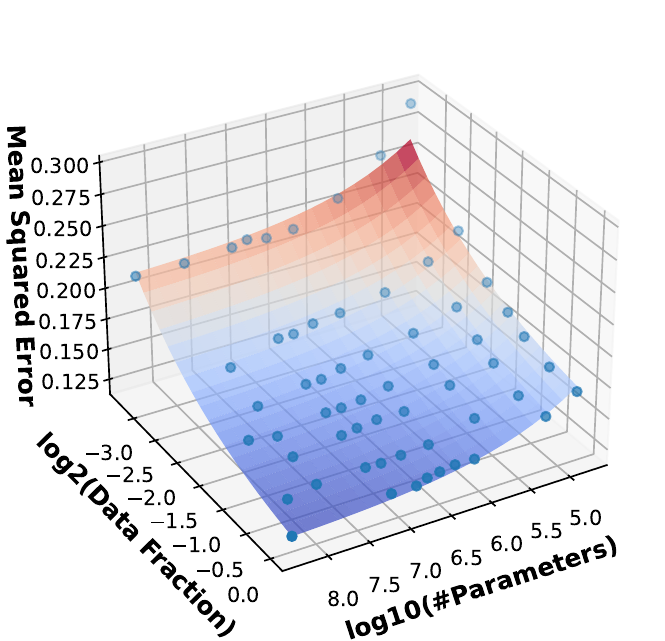}
    \end{minipage}
    }
    \subfigure[Scaling behaviors on \ppa{}, $R^2=0.86$.\label{subfig:ppa combined}]{
    \begin{minipage}[b]{0.45\textwidth}
    \includegraphics[width=0.9\columnwidth]{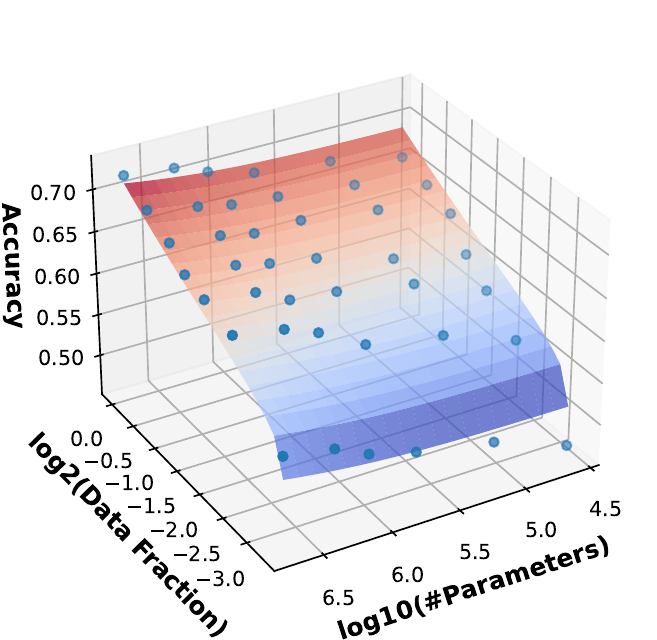}
    \end{minipage}
    }

    \caption{The comprehensive scaling behaviors of GIN on \PCQ{} and \ppa{} datasets. For \PCQ{}, the color corresponds to the value of mean square error; for \ppa{}, the color corresponds to accuracy. The surfaces predicted by the neural scaling law fit well with the empirical values on both datasets, as indicated by the high $R^2$ values.}\label{fig:combined}

\end{figure}

More scaling behaviors on the \ppa{} dataset are illustrated in Figure~\ref{subfig:ppa model} and~\ref{subfig:ppa data}.
The way of varying dataset sizes and model sizes is similar to that of \PCQ{}.
However, since the task on \ppa{} is graph classification, we use accuracy as the performance metric and apply Equation~\ref{equation:testscore} to fit the scaling behaviors.
Observations from Figure~\ref{subfig:ppa model} and \ref{subfig:ppa data} align with those from Figure~\ref{subfig:pcq model} and~\ref{subfig:pcq data}, suggesting that Equation~\ref{equation:testscore} can describe the scaling behavior for graph classification tasks.


On the basis of the validity of data scaling law and model scaling law individually, we further show that the two formulations can be combined into a comprehensive form, similar to 
\cite{rosenfeld2019constructive}. For regression, the form is:
\begin{equation}
    \mathbf{\epsilon = a_1D^{-b_2} + a_2N^{-b_2} + \epsilon_\infty} \label{equation: combined loss}
\end{equation}
For classification tasks, the comprehensive form is:
\begin{equation}
    \mathbf{s = -a_1(D+c_1)^{-b_1} - a_2(N+c_2)^{-b_2} + s_\infty}\label{equation: combined score}
\end{equation}
We verify the comprehensive forms by fitting the empirical values to Equation \ref{equation: combined loss} and \ref{equation: combined score}.
The experimental results are shown in Figure~\ref{fig:combined},
where the high $R^2$ values suggest that the model scaling and data scaling laws can be well unified by Equation~\ref{equation: combined loss} and \ref{equation: combined score}. 

\textbf{Observation 1. \textit{The model and data scaling behaviors in the graph domain can be described by neural scaling laws.}}

The above observations suggest that the scaling behaviors for graph property prediction tasks follow neural scaling laws and can be correctly predicted by Equation~\ref{equation:error} and \ref{equation:testscore}. Thus, it is possible to attain expected performance on graph tasks by enlarging the model and dataset size, exhibiting the potential of building large graph models. Despite the general success, there are still some unique chanllenges for scaling laws in the graph domain. We will discuss the challenges for the model scaling in Section~\ref{sec:The Effect of the number of model layers}, and those for the data scaling in Section~\ref{sec:DataScaling}.

\section{The Effect of Model Depth on Graph Scaling Laws\label{sec:The Effect of the number of model layers}} A unique question in the graph domain is whether the model depth would affect the model scaling law, or more specifically, the scaling law coefficients in Equation~\ref{equation:testscore}. 
Previous research~\cite{kaplan2020scaling} in the NLP domain finds that the model depth does not have effects on the scaling coefficients, i.e., the scaling curves for transformers of different numbers of model layers will overlap. 
Therefore, conducting preliminary experiments to determine the optimal number of layers for large language models (LLMs) is not required, which effectively narrows the hyperparameter search space.

\begin{figure*}[ht!]
    \centering
    \subfigure{
    \begin{minipage}[b]{0.23\textwidth}
    \includegraphics[width=\columnwidth]{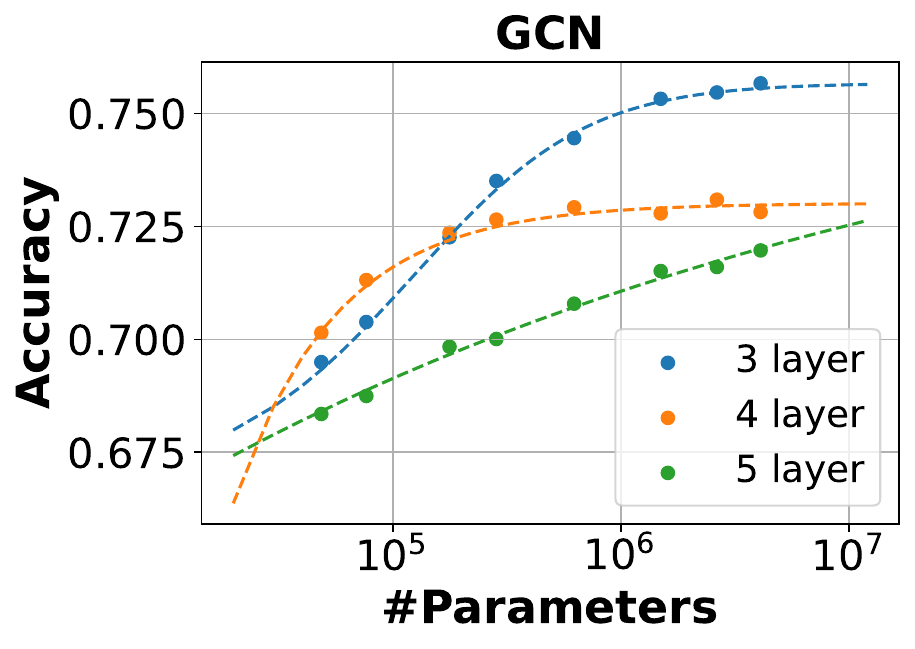}
    \end{minipage}
    }
    \subfigure{
    \begin{minipage}[b]{0.23\textwidth}
    \includegraphics[width=\columnwidth]{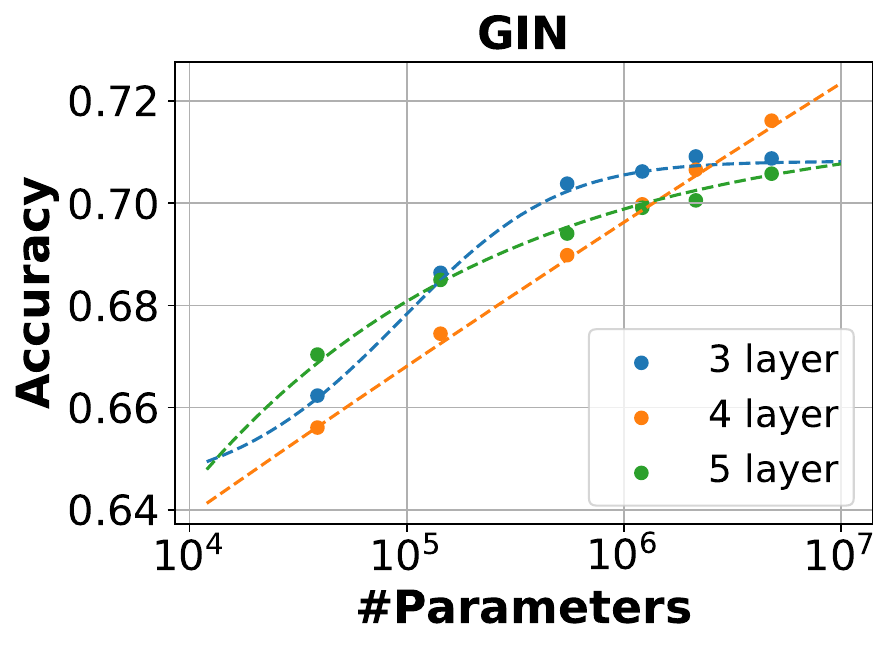}
    \end{minipage}
    }
    \subfigure{
    \begin{minipage}[b]{0.23\textwidth}
    \includegraphics[width=\columnwidth]{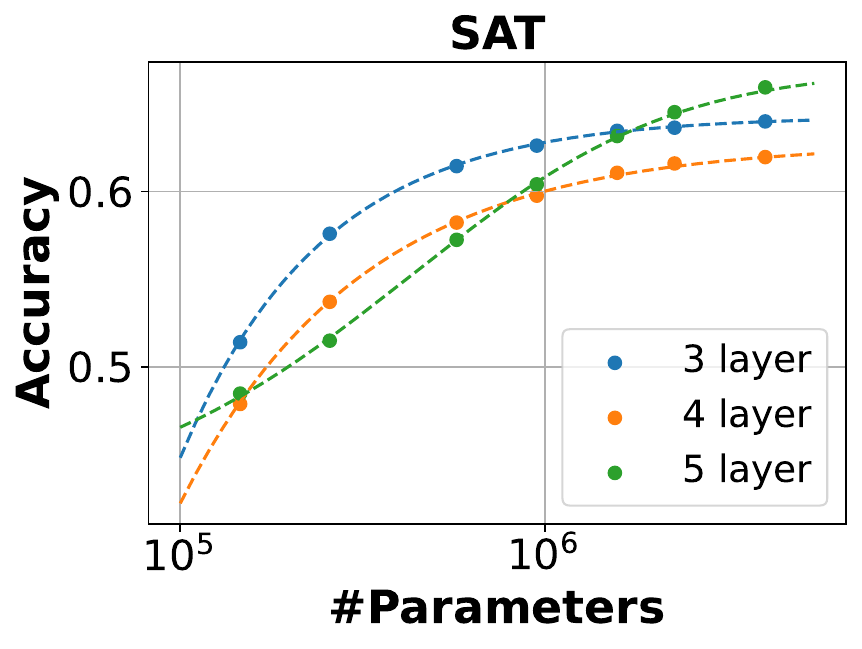}
    \end{minipage}
    }
    \subfigure{
    \begin{minipage}[b]{0.23\textwidth}
    \includegraphics[width=\columnwidth]{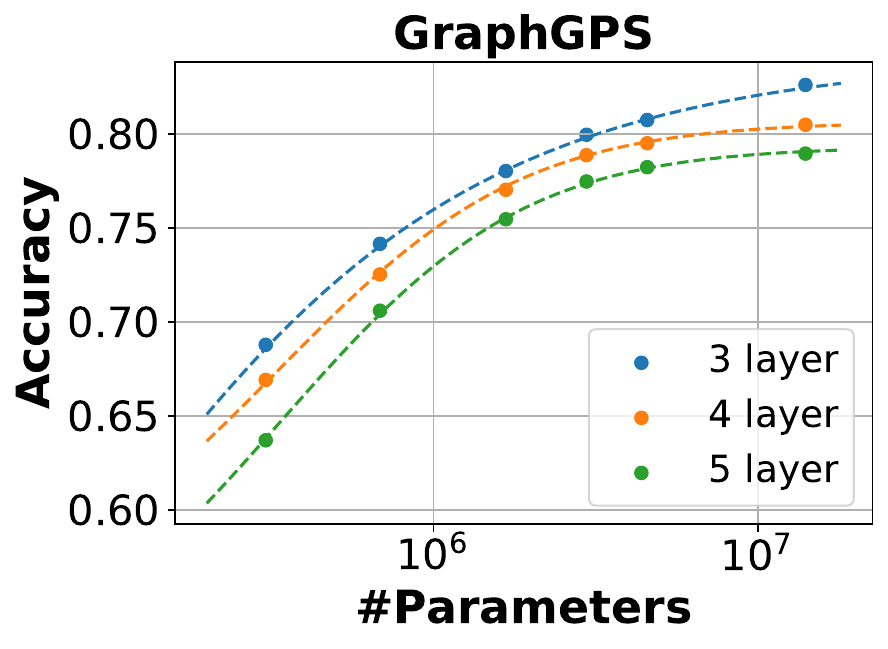}
    \end{minipage}
    }
    \caption{Model scaling behaviors of GNNs and graph transformers with varying model depths. The color corresponds to the model layer. All the models exhibit distinct scaling behaviors with different model depths.}\label{fig:layer gnn ppa}
\end{figure*}

Nonetheless, the deep graph models have different architectures, i.e., non-parametric aggregation layer, from vanilla transformers, which induces unique problems in graph learning such as over-smoothing~\cite{rusch2023survey} and over-squashing~\cite{di2023does}.
Thus, it is desired to study whether the conclusion from the NLP domain still holds on graphs.

\begin{wrapfigure}{R}{0.35\textwidth}
    \vspace{-1.9em}
    \centering
    \subfigure{
    \begin{minipage}[b]{0.28\textwidth}
    \includegraphics[width=\columnwidth]{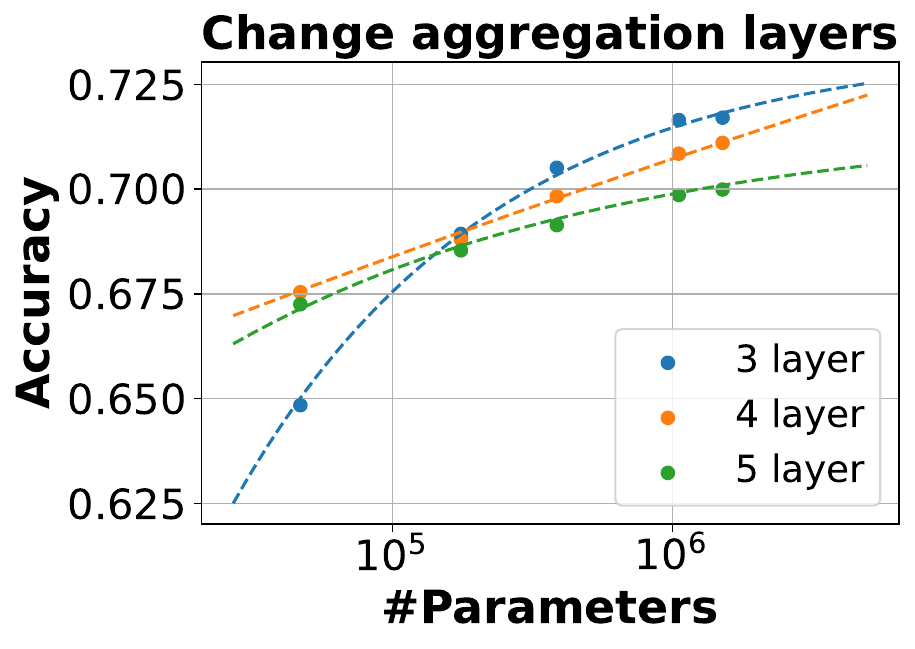}
    \end{minipage}
    }
    \subfigure{
    \begin{minipage}[b]{0.28\textwidth}
    \includegraphics[width=\columnwidth]{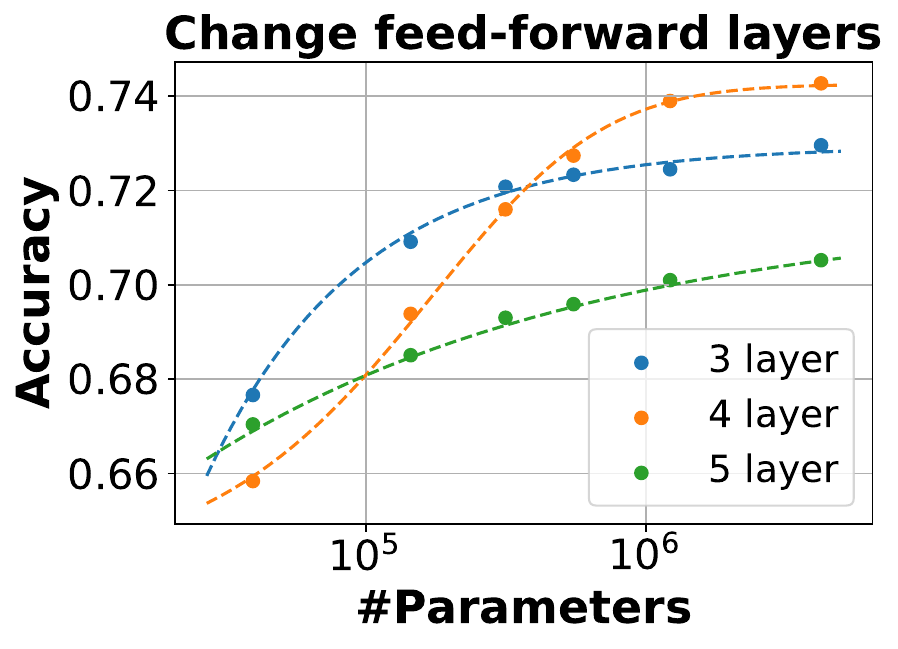}
    \end{minipage}
    }
    \vskip -0.7em
    \caption{Model scaling behaviors when fixing the number of aggregation layers or feed-forward layers. 
    \label{fig:layer decouple}}
    \vskip -1.0em
\end{wrapfigure}


To ensure the generalization of our investigations, we choose two representative GNNs~(GCN \& GIN) and two powerful graph transformers~(SAT \& GraphGPS).  
For these models, we choose the layer number in $[3,4,5]$ and fix it but change the model width to attain varying model sizes. 
All the models are trained on the whole \ppa{} training set and then evaluated for model performance. The model scaling curves of models with different model depths are presented in Figure~\ref{fig:layer gnn ppa}.
Observations can be made as follows:
(1) The scaling behaviors of models with different numbers of layers are distinct.
For each model, the shape of the scaling law curves would change when the model depth changes, exhibiting different increasing speeds and upper bounds for performance.
(2) The optimal number of layers for different models could be different. For GCN and GraphGPS, models with $3$ layers have the best performance, but for GIN the optimal layer number is $4$, and for SAT is $5$.
Similar observations on more datasets can be found in Appendix~\ref{app:model depth}.

\textbf{Observation 2. \textit{Deep graph models with varying depths will have different model scaling behaviors.}}

Current observations support that there is no one-size-fits-all model depth for existing graph learning models.
The optimal number of model layers could be task-specific and model-specific. 
We emphasize the necessity of preliminary experiments for the optimal number of model layers before scaling up the model parameters. 
Further, we look deep into the relationship between the non-parametric aggregation operation and the layer difference of graph model scaling. To achieve this goal, we first decouple the GIN layer into one feed-forward layer and one non-parametric aggregation layer, then repeat the above experiments by varying the number of feed-forward or aggregation layers while fixing the other to be 5. 
The results are shown in Figure~\ref{fig:layer decouple}. We observe that no matter whether fixing the aggregation layer or feed-forward layer, there will always be layer differences in model scaling.
These observations suggest that the non-parametric aggregation could be one of the reasons for the layer difference.
As for the observations when increasing the feed-forward layers, we speculate the local reception field of GNN (different from the global attention of vallina transformer) to be the reason.
Notably, we do not have any principle guidance helping to automatically select the best model layer so far and we still need to select the optimal layer number for specific graph learning tasks. Thus, building a new model whose model scaling behaviors are agnostic to the number of layers could be beneficial to the development of large graph models.


\section{Proper Data Metric of Graph Scaling Laws\label{sec:DataScaling}}

In this section, we look into the data scaling law on graphs. Specifically, we re-examine data metrics in previous studies and reveal the disadvantage of the number of graphs as the data metric in the data scaling law. For instance, a small molecule graph could have a large size difference compared with a social network.
Instead, we identify the total number of nodes or edges in the training set as a better metric with experimental evidence. 

In previous studies, the data metric in scaling laws is the number of graph samples, i.e., how many graphs in the whole training set~\cite{rosenfeld2019constructive, chen2023uncovering}.
Nonetheless, using the number of graphs as the metric neglects the size variability of graphs. 
For instance, a graph with $10$ edges and a graph with $1,000$ edges are treated equivalently when the number of graphs is the metric, even though they could differ remarkably in training costs and impacts on model performance improvement.
Empirically, we conduct two controlled experiments to show that the number of nodes or edges is the better data metric in general.

\begin{wrapfigure}{R}{0.35\textwidth}
    \vspace{-1.9em}
    \centering
    \subfigure{
    \begin{minipage}[b]{0.28\textwidth}
    \includegraphics[width=\columnwidth]{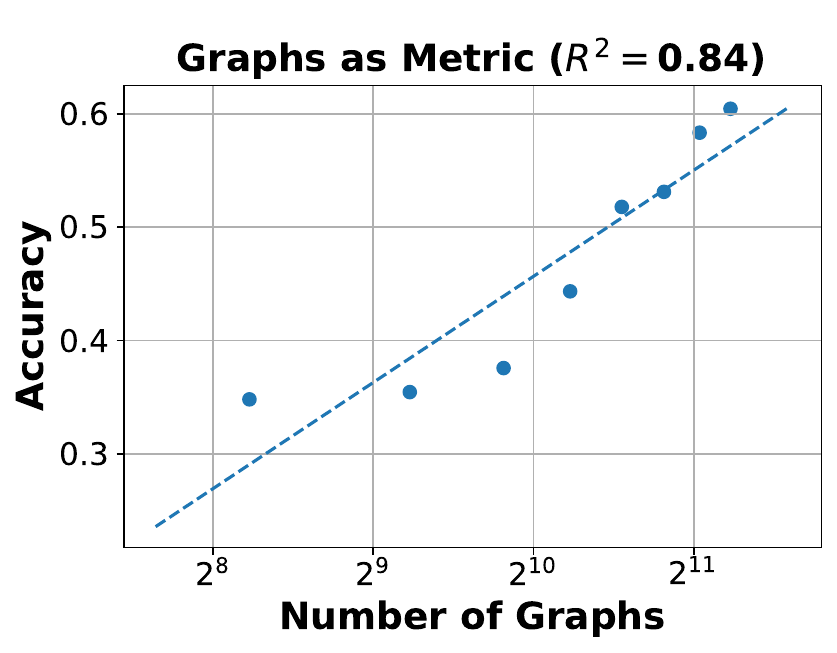}
    \end{minipage}
    }
    \subfigure{
    \begin{minipage}[b]{0.28\textwidth}
    \includegraphics[width=\columnwidth]{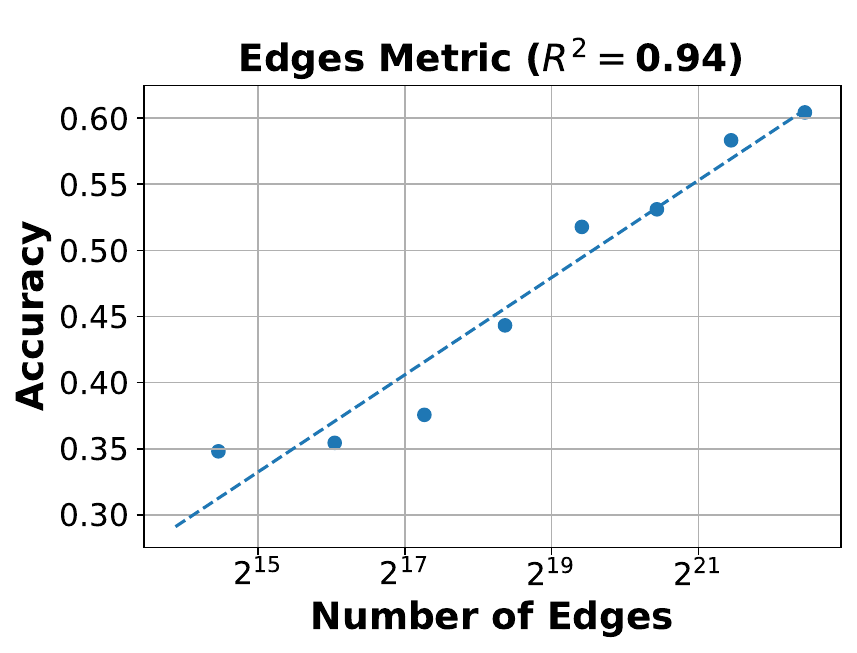}
    \end{minipage}
    }
    \vskip -0.7em
    \caption{Comparison between the number of graphs and edges as data metric 
    \label{fig:Sythetic}}
    \vskip -0.7em
\end{wrapfigure}


\paragraph{Experiment 1: From the data scaling perspective}
In this experiment, we show that the number of nodes (edges) as the data metric can provide a better fitting of scaling laws than the number of graphs. 
To highlight the contrast, the training set is scaled up in the following way: each time we add a subset with the same number of graphs but an increasing number of total nodes/edges. 
However, the graph size distribution often varies among classes of real-world datasets. To eliminate the bias, we adopt the setup in ~\cite{ying2019gnnexplainer} and construct a synthetic dataset with a motif recognition task. 
Notably, in the setup~\cite{ying2019gnnexplainer}, the synthetic graphs are generated as Barabasi Albert graphs, whose nodes are connected with a fixed number of edges.
Thus, we can view the numbers of nodes and edges as equivalent in this setting.
Specifically, the dataset includes three classes, i.e., three motifs~(Cycle, Grid, House) to recognize. 
We generate 8 subgroups for each class, where each subgroup contains 100 graphs with the same number of nodes $i$~($i \in [25, 50, 100, 200, 400, 800, 1600, 3200]$). 
Each time to increase the training set size, we add the subgroups of three classes with the same number of nodes $i$ ($i$ increases monotonically from 25 to 3200). Then we can ensure that the total number of graphs from each class is balanced. 
After recording the model performance of varying training set sizes, we fit the data scaling law to the empirical values with the number of nodes (edges) as the data metric repectively. 
The results of the number of edges are shown in Figure~\ref{fig:Sythetic}.
Results with the number of nodes as the metric can be found in Appendix~\ref{app:results with node metric}. 
From the figure, we observe that the number of nodes as the data metric gives a more accurate fitting with a higher $R^2$ value, which supports our argument.

\begin{wrapfigure}{R}{0.35\textwidth}
  
    \includegraphics[width=0.3\columnwidth]{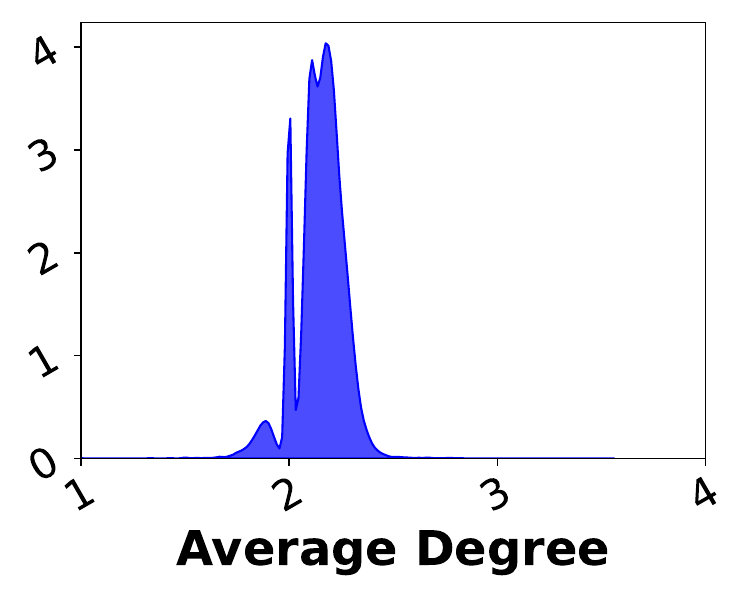}
 \vskip -1.0em
    \caption{Average degree distribution of \molhiv{} dataset.
    \label{fig:hiv degree}}
        \vskip -1.0em
\end{wrapfigure}
\begin{figure*}[ht!]
    \centering
    \subfigure[\label{subfig:equal_graph_number_Molhiv.pdf}]{
    \begin{minipage}[b]{0.45\textwidth}
    \includegraphics[width=\columnwidth]{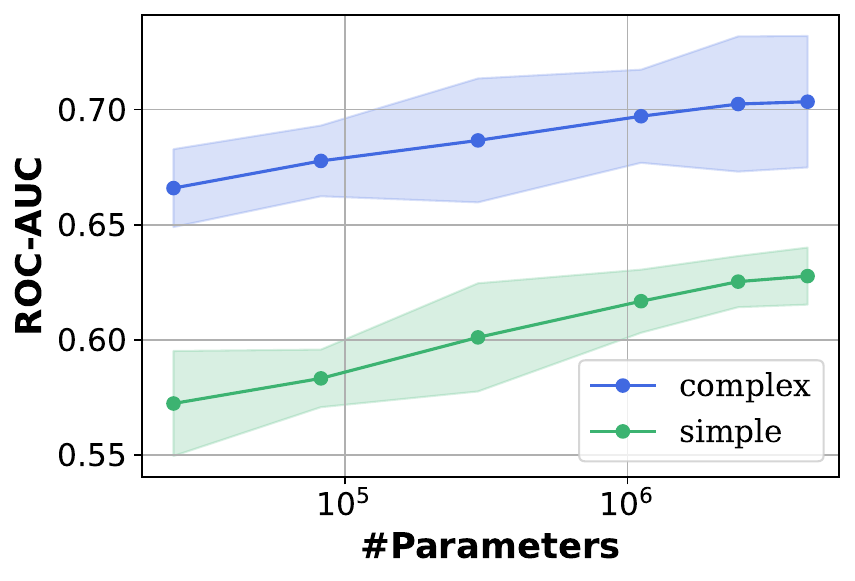}
    \end{minipage}
    }
    \subfigure[\label{subfig:equal_edge_number_Molhiv.pdf}]{
    \begin{minipage}[b]{0.45\textwidth}
    \includegraphics[width=\columnwidth]{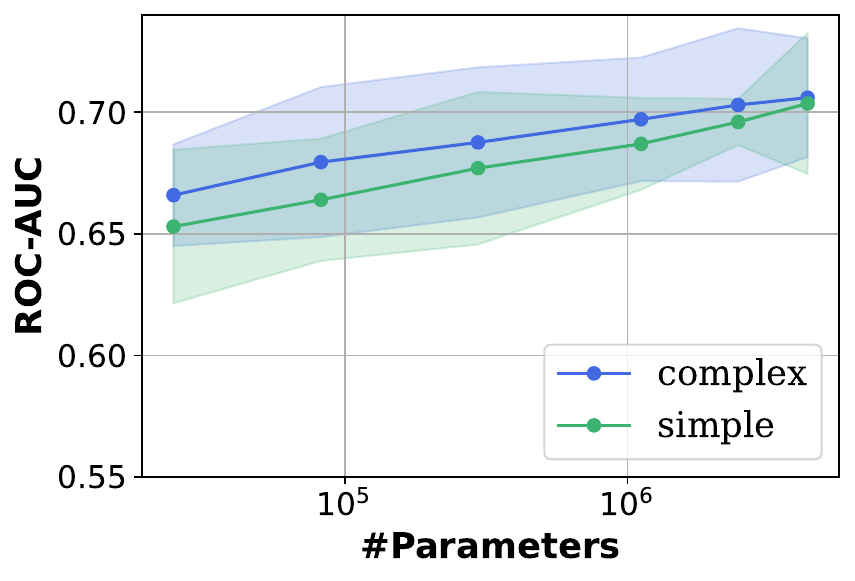}
    \end{minipage}
    }
    \caption{Model scaling curves of GIN on two training subsets of different sizes. (a) Two subsets have the same total number of graphs but different total number of edges. (b) Two subsets have the same total number of edges but different total number of graphs. The shade corresponds to the performance variance of repeated experiments.}\label{fig:graph number bad}
\end{figure*}

\paragraph{Experiment 2: From the model scaling perspective}
For real-world datasets, we contrast the two metrics by comparing the model scaling curves of the same model on two training sets with the same number of graphs but different number of nodes/edges.
Since this study focused on in-domain graph learning, the graph samples tend to have similar average degrees, as shown in Figure~\ref{fig:hiv degree}. 
The degree distributions of other graph datasets have the same trend, which can be found in Appendix~\ref{app:averge degree}.
Hence, we could treat the number of nodes and edges as nearly equivalent metrics.
We discuss the situations with the number of edges as the metric in this subsection and results with the number of nodes can be found in Appendix~\ref{app:results with node metric}. 
We first divide the whole training sets of one dataset into two subsets as follows: 
(1) We first arrange all graph samples of class $i$ into a list $\mathcal{G}_i$ in an ascending order of their numbers of edges. 
(2) We choose the first half of all $\mathcal{G}_i$ as the ``simple'' training subset, and then the second half as the ``complex'' subset. 
Thus, these two subsets have the same number of graphs while the graphs in the complex subset have more edges.
The validation and test sets remain as the original.
Then we train GIN of varying sizes on these two subsets and report the performance on \molhiv{} in Figure~\ref{fig:graph number bad}. 

If the number of graphs can accurately metric the data volume, the curves of simple and complex subsets in Figure~\ref{subfig:equal_graph_number_Molhiv.pdf} should have small differences or even overlap with each other. However, their performance curves exhibit considerable differences, indicating that the two subsets contain different amounts of data. 
For comparison, we attain the two subsets in another way: the simple subset has the first $m$ samples in $\mathcal{G}_i$, and the complex subset has the remaining $n$ samples, where the two subsets contain the same number of edges. It is easy to validate that $m>n$. 
Then we train models of varying sizes on these two subsets and report their performance on \molhiv{} in Figure~\ref{subfig:equal_graph_number_Molhiv.pdf}.

Compared to Figure~\ref{subfig:equal_graph_number_Molhiv.pdf}, the curves in Figure~\ref{subfig:equal_edge_number_Molhiv.pdf} have much smaller differences, which indicates that the total number of edges could be a better metric than the number of graphs. 

\textbf{Observation 3. \textit{The number of edges is a better data metric for data scaling law compared to the number of graphs.}}

In addition, using the total number of edges as the data metric in scaling laws has two more advantages:
(1) Knowing the GNN architecture and the number of edges in a dataset, the compute cost of training can be directly calculated, which enables us to build connections between data scaling law and compute scaling law as in \cite{kaplan2020scaling}. The detailed explanation is in Appendix~\ref{app:edge and flops}.
(2) Furthermore, using the number of edges as the data metric potentially provides a unified way to describe the data scaling behaviors of node classification and link prediction tasks, which will be thoroughly studied in the following section.

\section{Results on Node Classification and Link Prediction\label{sec:Results on Node Classification and Link Prediction}} 
\begin{figure*}[ht!]
    \centering
    \subfigure{
    
    \begin{minipage}[b]{0.3\textwidth}
    \includegraphics[width=\columnwidth]{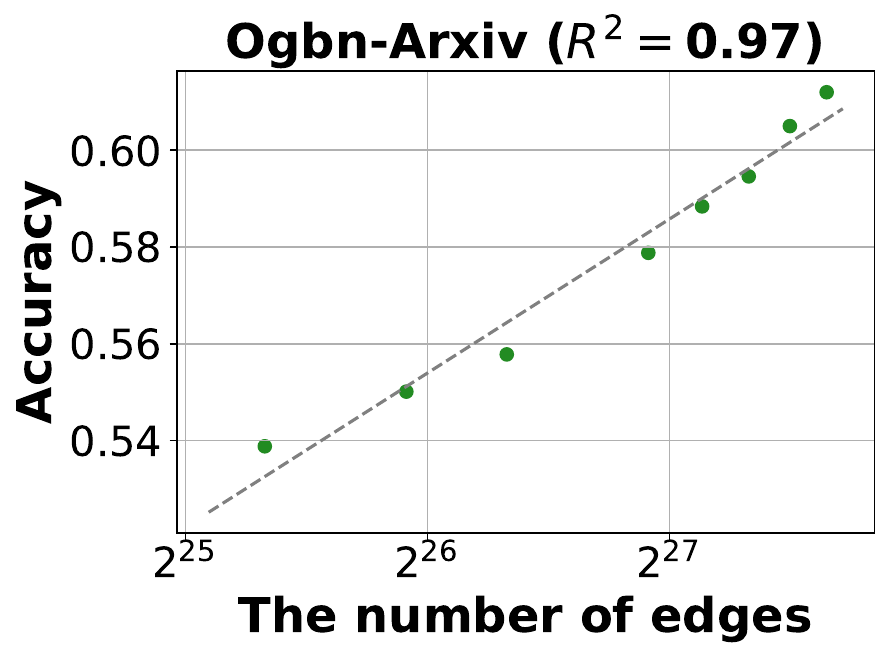}
    \end{minipage}
    }
    \subfigure{
    
    \begin{minipage}[b]{0.3\textwidth}
    \includegraphics[width=\columnwidth]{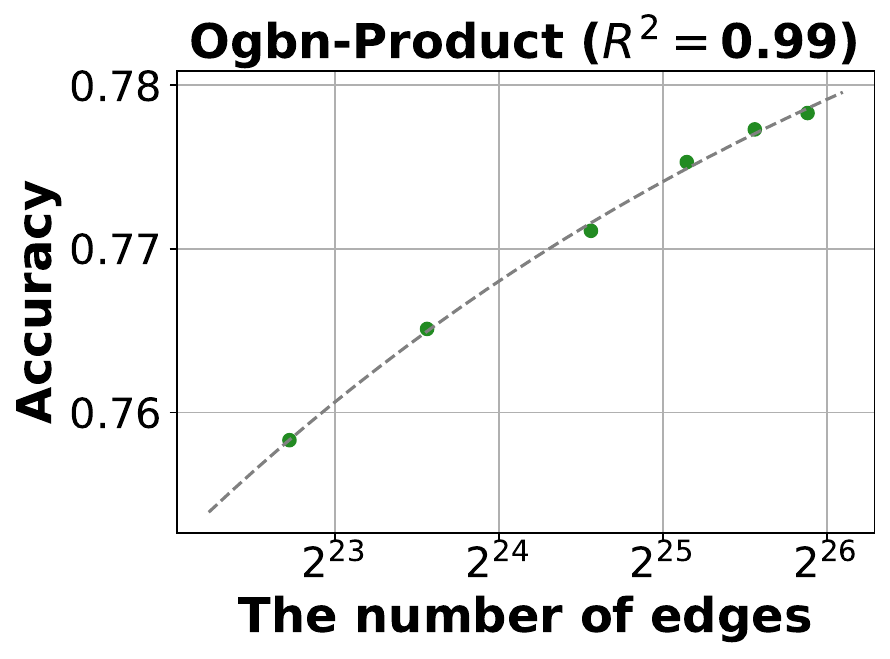}
    \end{minipage}
    }
        \subfigure{
    
    \begin{minipage}[b]{0.3\textwidth}
    \includegraphics[width=\columnwidth]{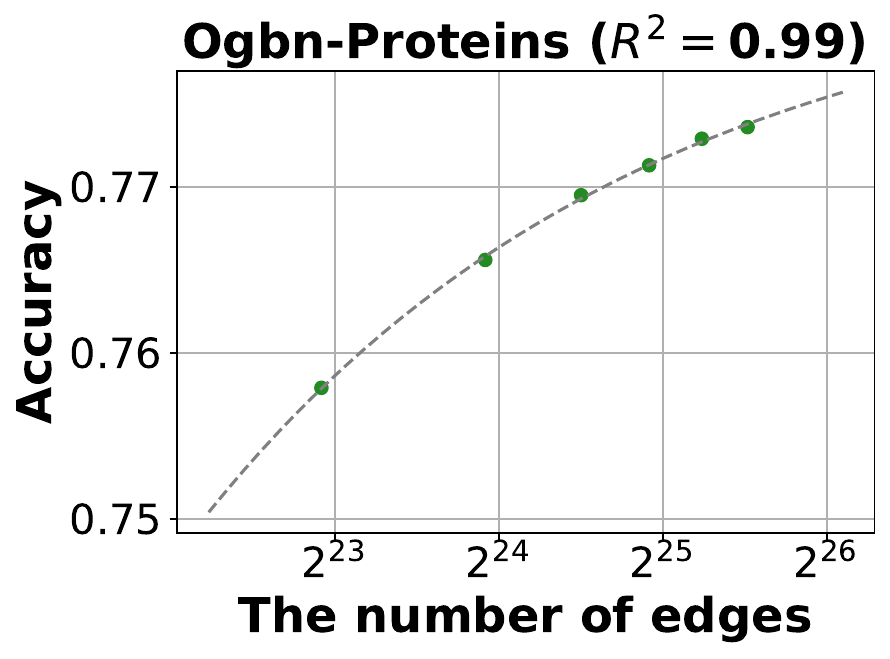}
    \end{minipage}
    }
    \vfill
    \subfigure{
    
    \begin{minipage}[b]{0.3\textwidth}
    \includegraphics[width=\columnwidth]{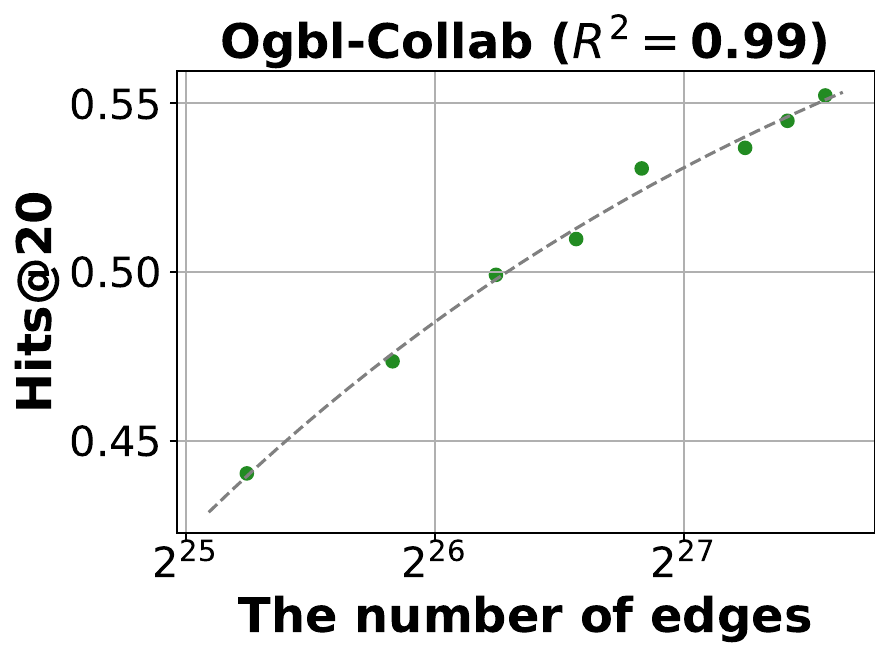}
    \end{minipage}
    }
        \subfigure{
    
    \begin{minipage}[b]{0.3\textwidth}
    \includegraphics[width=\columnwidth]{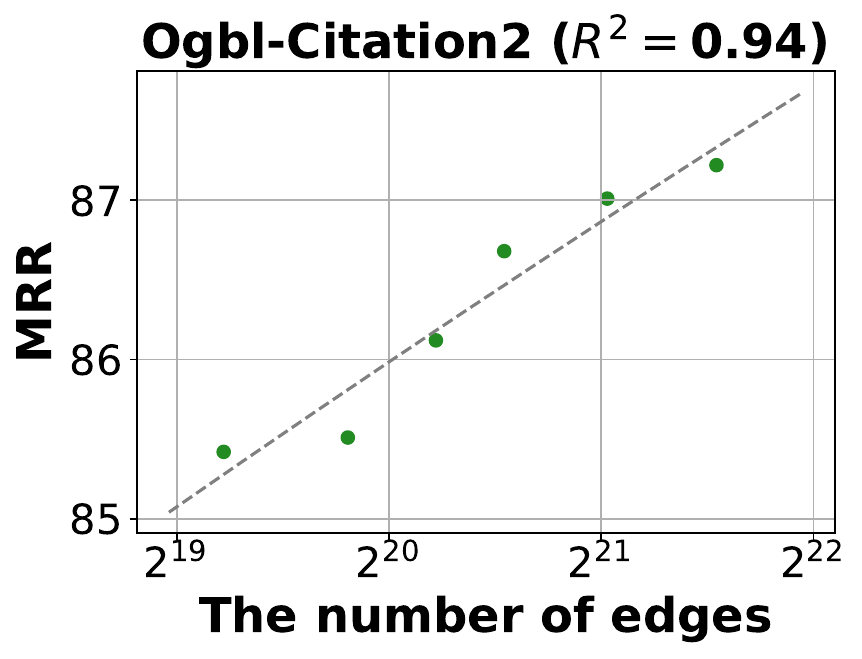}
    \end{minipage}
    }
            \subfigure{
    
    \begin{minipage}[b]{0.3\textwidth}
    \includegraphics[width=\columnwidth]{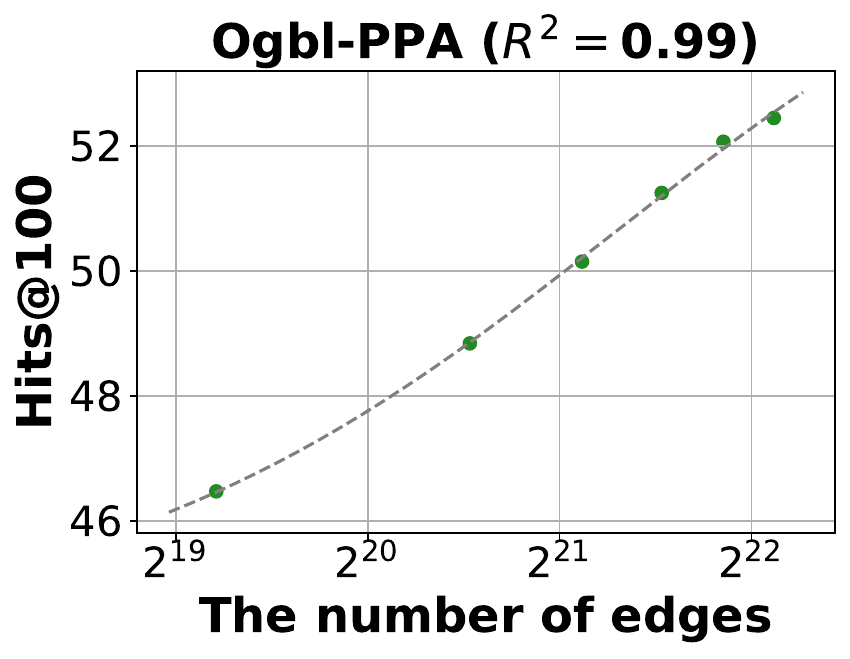}
    \end{minipage}
    }

    \caption{Both node classification and link prediction follow the data scaling law with the total edge number as the data metric.}\label{fig:node and link}

\end{figure*}

Furthermore, the data scaling law with the number of nodes~(edges) as the data metric can be naturally extended to node classification and link prediction tasks, potentially providing a unified view of data scaling behaviors in the graph domain. 
There is only one graph in most of the node classification and link prediction datasets, thus it is infeasible to use the number of graphs to measure the training data volume. 
Nonetheless, the total number of nodes~(edges) is able to serve as the data metric for these tasks. 
Moreover, with the total number of nodes~(edges) as the data metric, we can potentially attain a unified data scaling law on graphs by directly extending Equation~\ref{equation:testscore} to node classification and link prediction tasks.
To verify this, we conduct experiments on the data scaling behaviors on both node classification and link prediction tasks.
Here we only provide results with the number of edges as metric, for the results with the number of edges as metric, please refer to Appendix~\ref{app:node and link with node metric}.
For node classification, we train a $3$-layer GraphSAGE~\cite{hamilton2017inductive} on \arxiv{}, \products{} and \proteins{}, where the hidden vector size is set to $128$.
For link prediction, we train a $3$-layer SEAL~\cite{SEAL} on \collab{}, \linkppa{} and \cit{}, where the model takes one-hop neighbors of the target node pairs as input and its hidden vector size is set to $256$. 
We take the total number of edges on which models are trained as the training set size for both tasks. 
The results are demonstrated in Figure~\ref{fig:node and link}. Similar to graph classification tasks,  Equation~\ref{equation:testscore} fits well to the data scaling behaviors, as evidenced by the high $R^2$ values. 

\textbf{Observation 5. \textit{Node classification, link prediction, and graph classification tasks could follow the same form of data scaling law with the number of nodes~(edges) as data metric.}}

\section{Conclusions\label{sec:Conclusions}}
In this work, we take the first step to investigate neural scaling laws on graphs.
In particular, we focus on two basic forms of neural scaling laws: the model scaling law and the data scaling law. 
We first verify the formulations that can describe both scaling behaviors in the graph domain.
For the model scaling law, we identify and investigate some graph-specific phenomena, revealing different observations from CV and NLP domains.
For the data scaling law, we reform it by adopting the number of edges as the data metric and extend the reformed version to node classification and link prediction tasks.
With more attention has been attracted to the graph foundation models~\cite{mao2024graph,liu2023one,huang2024prodigy, galkin2023towards}, our findings shed light on the scaling behaviors of deep graph models and can provide valuable instructions for developing large graph models.

\textbf{Limitations.} Next, we discuss the limitations of this work, which could be further explored as future works. 
(1) There is no large graph dataset as large as CV or NLP datasets. We conduct our experiments on the largest graph datasets that we can attain in academics. 
Even though, this initial study is still meaningful. 
Only after the data scaling law on graphs has been verified, more efforts would be made to construct larger graph datasets, and the experience from this work could also be used to guide the process.
For instance, Tang et al~\cite{Wenzhuo} adapt the data scaling scaling perspective and improve the crosse-domain graph learning performance by generating synthetic graph data for training by diffusion.
(2) Since there has not been a well-established pretraining task for graph learning, we limit our investigation to the supervised learning on graphs.
(3) In our exploration, we assume that all node~(edge) features are in a unified feature space. 
We do not consider heterogeneous feature problems in this work.

\bibliographystyle{unsrtnat}
\bibliography{reference}
\newpage
\appendix
\section{Emperiment Settings.\label{app:experiment settings}}
Here we introduce the training datasets used and training hyperparameters in the experiments. 

\textbf{Datasets.} The basic situations of the datasets we use are listed in the Table~\ref{tab:graph level datasets},~\ref{tab:node datasets} and ~\ref{tab:link datasets} below.

\begin{table}[h]
\small
\centering
 \caption{Graph Property Prediction Datasets Statistics.} \label{tab:graph level datasets}
 \vspace{0.5em}
\begin{tabular}{cccccc}
\toprule
 & \PCQ & \ppa{} & \molhiv{} & \molpcba{} & \reddit{}  \\
 \midrule

Domains & \multicolumn{1}{r}{Molecule} &\multicolumn{1}{r}{Protein} & \multicolumn{1}{r}{Molecule} &\multicolumn{1}{r}{Molecule}&\multicolumn{1}{r}{Social}\\
\#Graphs & \multicolumn{1}{r}{3,746,619} & \multicolumn{1}{r}{158,100} & \multicolumn{1}{r}{41,127} & \multicolumn{1}{r}{437,929}&\multicolumn{1}{r}{11,929}\\
\#Total Nodes & \multicolumn{1}{r}{52,970,652} & \multicolumn{1}{r}{38,481,540} & \multicolumn{1}{r}{1,048,739} & \multicolumn{1}{r}{11,386,154} & \multicolumn{1}{r}{4,669,130}\\
\#Total Edges & \multicolumn{1}{r}{54,546,813} & \multicolumn{1}{r}{358,270,410} & \multicolumn{1}{r}{1,130,993} & \multicolumn{1}{r}{12,305,805}&\multicolumn{1}{r}{5,450,241}\\
Tasks & \multicolumn{1}{r}{Regression} &\multicolumn{1}{r}{Classification} & \multicolumn{1}{r}{Classification} &\multicolumn{1}{r}{Classification}&\multicolumn{1}{r}{Classification}\\
Metrics& \multicolumn{1}{r}{MSE}&\multicolumn{1}{r}{Accuracy} &\multicolumn{1}{r}{ROC-AUC} &\multicolumn{1}{r}{AP}&\multicolumn{1}{r}{Accuracy}\\
\bottomrule
\end{tabular}

\end{table}

\begin{table}[h]
\small
\centering
 \caption{Node Classification Datasets Statistics. } \label{tab:node datasets}
 \vspace{0.5em}
\begin{tabular}{cccc}
\toprule
 & \arxiv{} & \products{} & \proteins{} \\
 \midrule

Domains & \multicolumn{1}{r}{Citation}  &\multicolumn{1}{r}{E-commerce} &\multicolumn{1}{r}{Biology} \\
\#Nodes & \multicolumn{1}{r}{169,343} & \multicolumn{1}{r}{2,449,029} & \multicolumn{1}{r}{132,534}\\
\#Edges & \multicolumn{1}{r}{1,166,243}  & \multicolumn{1}{r}{61,859,140} & \multicolumn{1}{r}{39,561,252} \\
Metrics& \multicolumn{1}{r}{Accuracy}&\multicolumn{1}{r}{Accuracy}&\multicolumn{1}{r}{ROC-AUC}\\

\bottomrule
\end{tabular}
\label{table:social_data}
\end{table}

\begin{table}[h]
\small
\centering
 \caption{Link Classification Datasets Statistics. } \label{tab:link datasets}
 \vspace{0.5em}
\begin{tabular}{cccc}
\toprule
 & \linkppa{} & \collab{} & \cit{} \\
 \midrule

Domains & \multicolumn{1}{r}{Citation}  &\multicolumn{1}{r}{E-commerce} &\multicolumn{1}{r}{Biology} \\
\#Nodes & \multicolumn{1}{r}{576,289} & \multicolumn{1}{r}{235,868} & \multicolumn{1}{r}{2,927,963}\\
\#Edges & \multicolumn{1}{r}{30,326,273}  & \multicolumn{1}{r}{1,285,465} & \multicolumn{1}{r}{30,561,187} \\
Metrics& \multicolumn{1}{r}{Hits@100}&\multicolumn{1}{r}{Hist@20}&\multicolumn{1}{r}{MRR}\\

\bottomrule
\end{tabular}
\label{table:social_data}
\end{table}

\textbf{Training Details.} For GIN and GCN models, we set the initial learning rate to 0.001 and use Adam optimizer~\cite{kingma2014adam} in the training process. For SAT and GraphGPS, we directly adopt the hyperparameters and optimizers that the authors present in their GitHub repos. For all the models, we opt against the learning rate with a scheduler. For \ppa{}, \arxiv{}, \products{}, \proteins{}, \linkppa{}, \cit{} and \collab{} datasets, the training epoch number is 50. For \PCQ{}, \molhiv, \molpcba{}, and \reddit{} datasets, the training epoch number is 100. We choose the test performance with the best corresponding validation performance as the final performance of the model.

\newpage
\section{More Results on the Validity of the Neural Scaling Law on Graphs\label{app:validity}}
Here we show the scaling behaviors of \textbf{GCN} on the \PCQ{} and \ppa{} datasets. All the empirical values fit well with the neural scaling law.

\begin{figure*}[ht!]
    \centering
    \subfigure[Model scaling on \PCQ{}, $R^2=0.98$.]{
    
    \begin{minipage}[b]{0.45\textwidth}
    \includegraphics[width=\columnwidth]{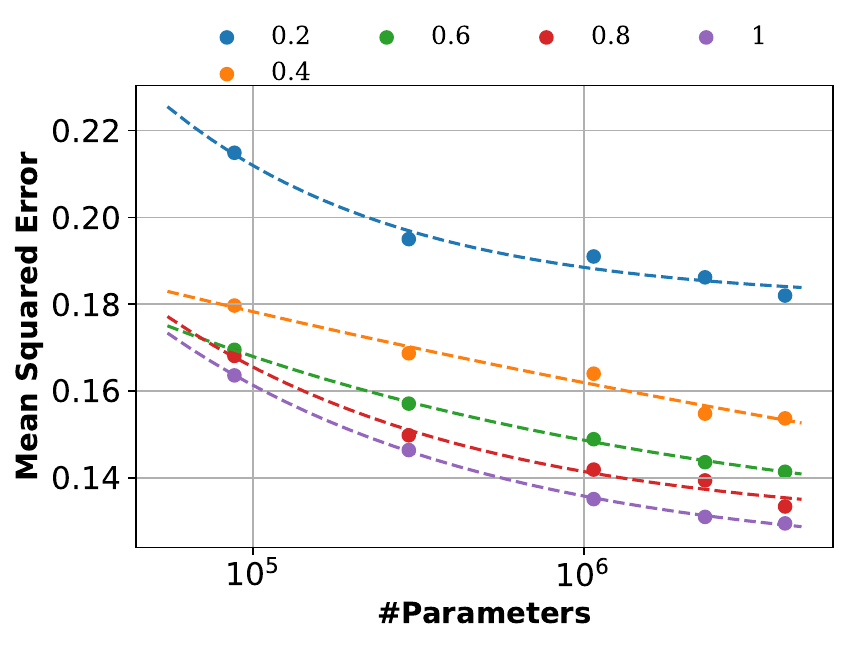}
    \end{minipage}
    }
    \subfigure[Data scaling on \PCQ{}, $R^2=0.98$.]{
    \begin{minipage}[b]{0.45\textwidth}
    \includegraphics[width=\columnwidth]{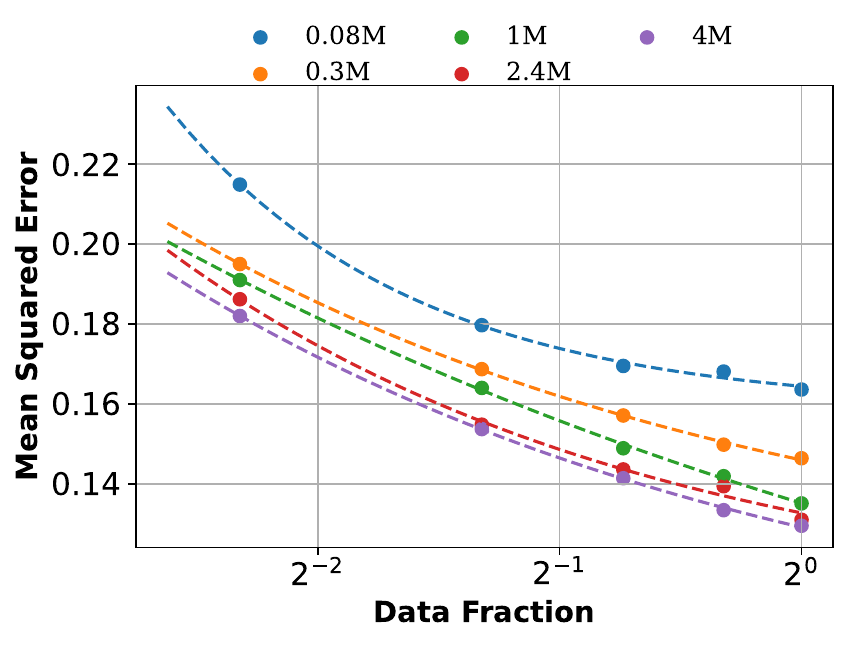}
    \end{minipage}
    }

    \caption{The model and data scaling behaviors of GCN on \PCQ{}. For model scaling, the color corresponds to the training data fraction; for data scaling, the color corresponds to the number of model parameters. The lower mean square error is, the better model performance is. The high $R^2$ values indicate that the scaling law curves fit well with the empirical values.}
  
\end{figure*}

\begin{figure*}[h]
    \centering
    \subfigure[Model scaling on \PCQ{}, $R^2=0.98$.]{
    
    \begin{minipage}[b]{0.4\textwidth}
    \includegraphics[width=\columnwidth]{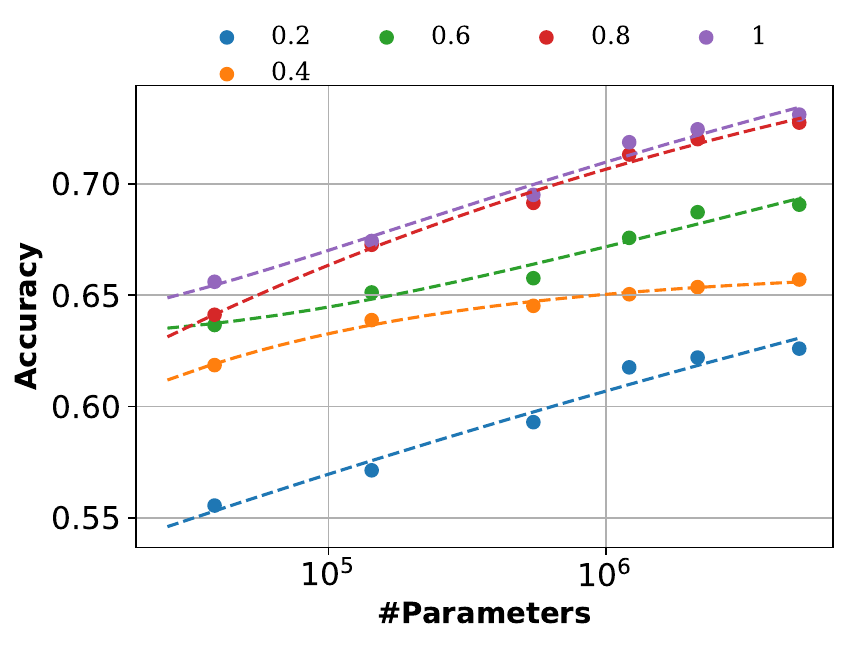}
    \end{minipage}
    }
    \subfigure[Data scaling on \PCQ{}, $R^2=0.99$.]{
    \begin{minipage}[b]{0.4\textwidth}
    \includegraphics[width=\columnwidth]{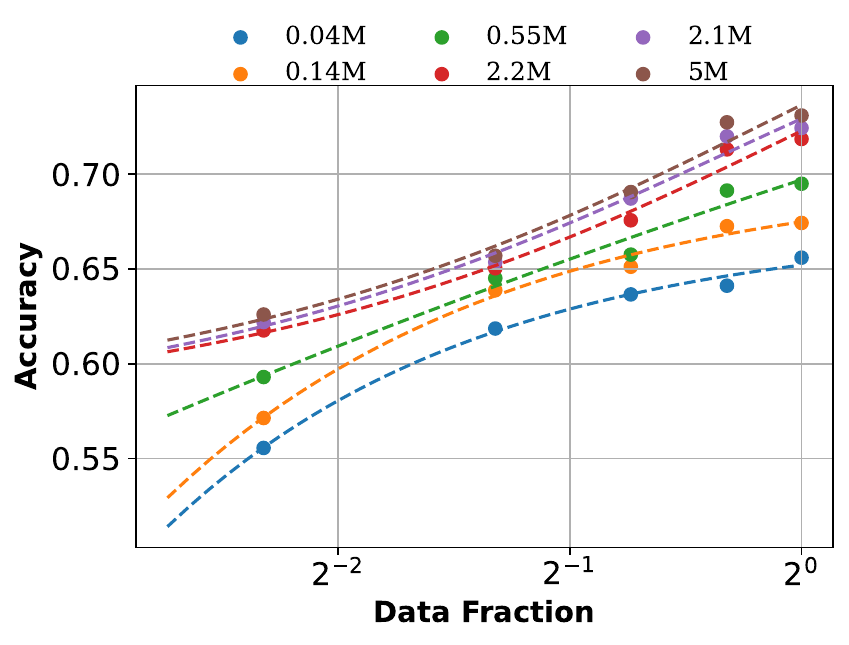}
    \end{minipage}
    }

    \caption{The model and data scaling behaviors of GCN on \ppa{}. For model scaling, the color corresponds to the training data fraction; for data scaling, the color corresponds to the number of model parameters. The higher the performance is, the better the model performance is. The high $R^2$ values indicate that the scaling law curves fit well with the empirical values.}

\end{figure*}

\begin{figure*}[h!]
    \centering
    \subfigure[Scaling behaviors on \PCQ{}, $R^2=0.97$.]{
    \begin{minipage}[b]{0.45\textwidth}
    \includegraphics[width=0.9\columnwidth]{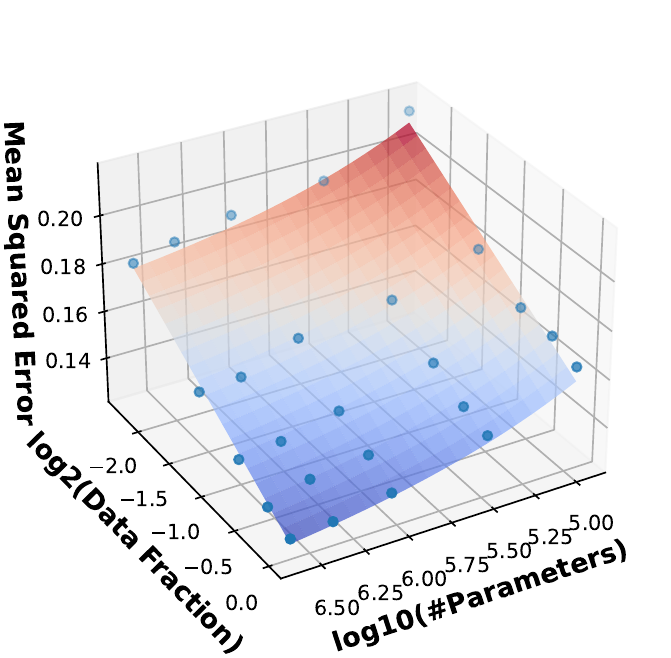}
    \end{minipage}
    }
    \subfigure[Scaling behaviors on \ppa{}, $R^2=0.97$.]{
    \begin{minipage}[b]{0.45\textwidth}
    \includegraphics[width=0.9\columnwidth]{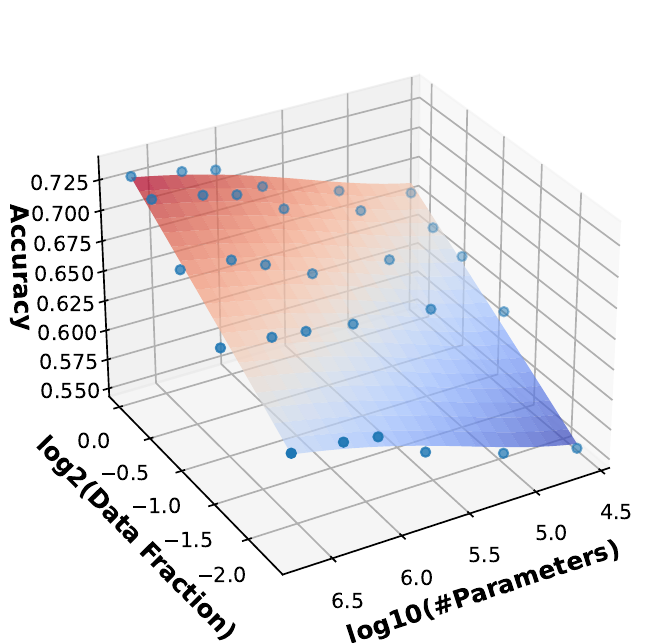}
    \end{minipage}
    }

    \caption{The comprehensive scaling behaviors of GCN on \PCQ{} and \ppa{} datasets. For \PCQ{}, the color corresponds to the value of mean square error; for \ppa{}, the color corresponds to accuracy. The surfaces predicted by the neural scaling law fit well with the empirical values on both datasets, as indicated by the high $R^2$ values.}
\end{figure*}
\section{Model Scaling Law on Limited Graph Data\label{app:Model scaling collapse}} 
\begin{figure*}[ht!]
    \centering
    \subfigure[Model scaling on \molpcba{}\label{subfig:collapse on Molpcba}]{
    \begin{minipage}[b]{0.3\textwidth}
    \includegraphics[width=\columnwidth]{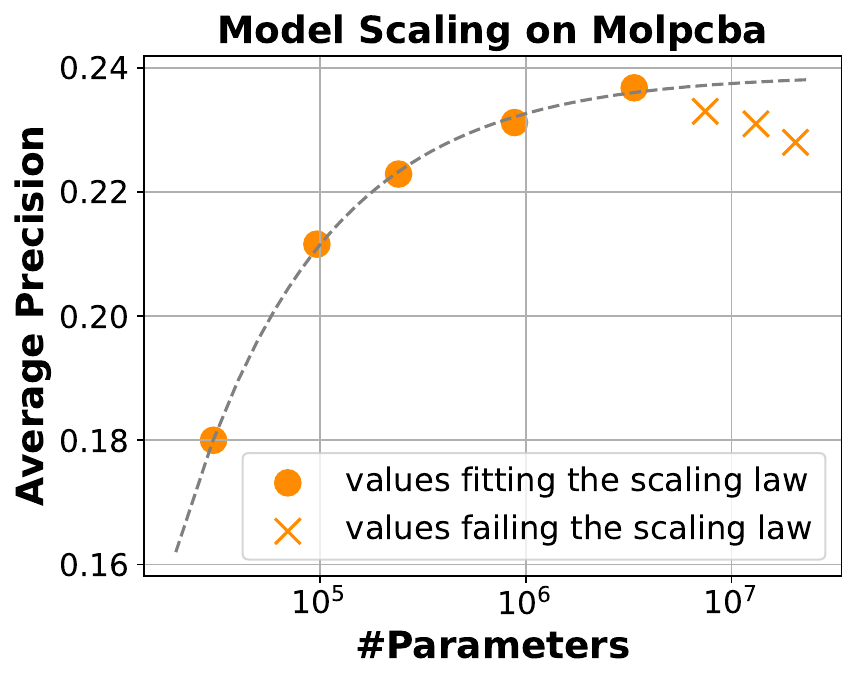}
    \end{minipage}
    }
    \subfigure[Training losses of models\label{subfig:training loss on Molpcba}]{
    \begin{minipage}[b]{0.3\textwidth}
    \includegraphics[width=\columnwidth]{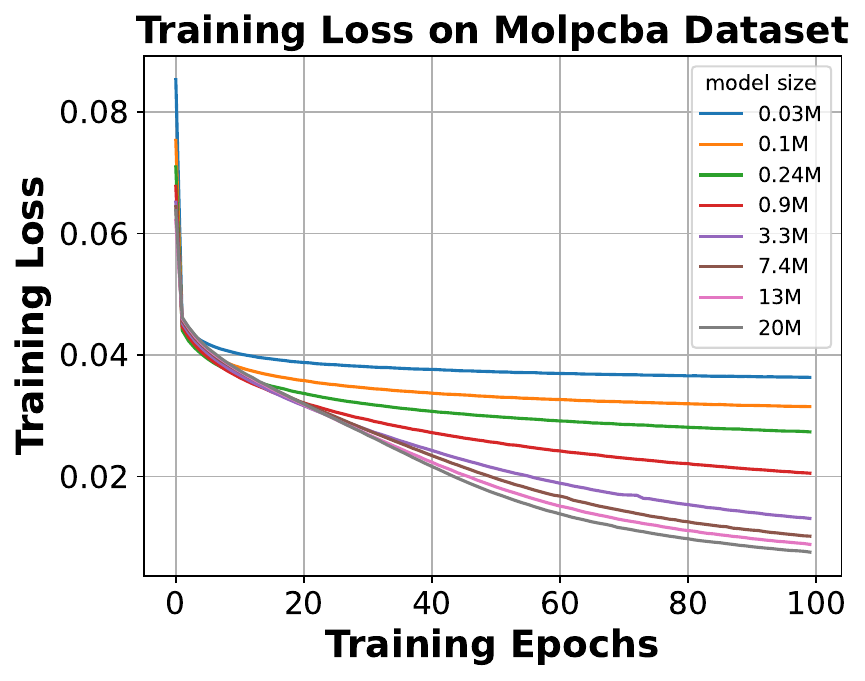}
    \end{minipage}
    }
    \subfigure[Validation losses of models\label{subfig:validation loss on Molpcba}]{
    \begin{minipage}[b]{0.307\textwidth}
    \includegraphics[width=\columnwidth]{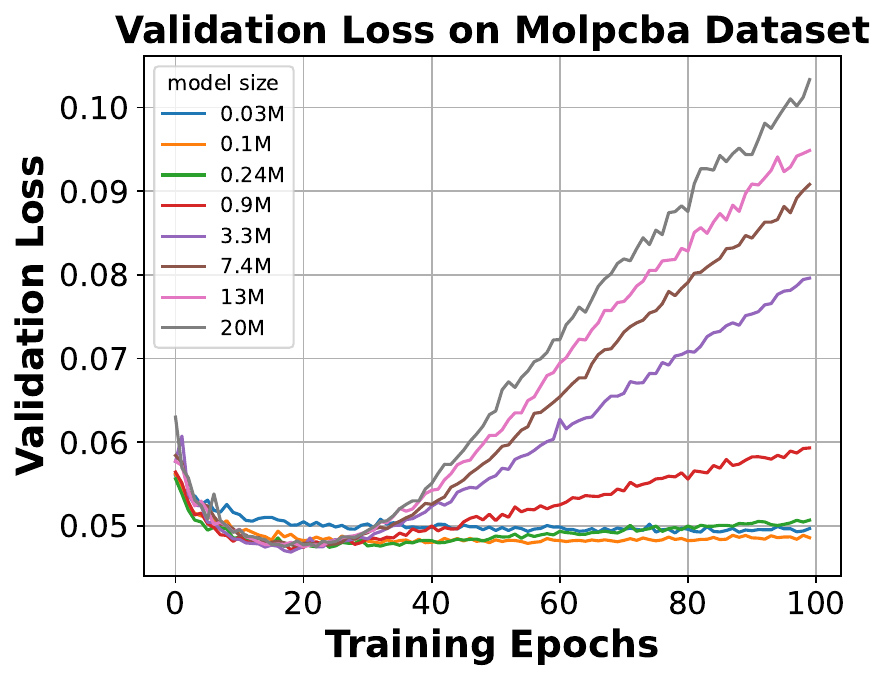}
    \end{minipage}
    }
    \caption{(a) The empirical performance of larger models falls below the prediction of model scaling law. (b)(c) The training losses and validation losses of models with varying sizes. The color corresponds to the number of model parameters. The model with larger sizes suffers more serious overfitting.}\label{fig:collapse pcba}
\end{figure*}

Despite the overall success, the scaling law could fail in certain situations. In particular, we find that the model scaling law could fail when the model size is too large relative to the training set size, causing the phenomenon called \textit{model scaling collapse}.  
To illustrate this phenomenon, we consider an example of scaling up 5-layer GIN on \molpcba{} dataset.
Similar to the setting in Section~\ref{sec:Basic Laws}, we vary the model size by changing the model width while fixing the training set size. 
Figure~\ref{subfig:collapse on Molpcba} displays the empirical accuracy of models across sizes alongside the model scaling curve.
At small model sizes, the accuracy values fit with the model scaling curve well. 
However, when the number of parameters is around $10^7$, the empirical values start dropping. 
Thus, the collapse of performance would make the predictions of the scaling law fail, which is referred to as model scaling collapse.

The model scaling collapse poses difficulty in building large graph models following the model scaling law. Thus, it is essential to explore why the collapse happens and how to mitigate it. 
In this work, we conjecture that one of the potential reasons for the scaling law collapse is overfitting. 
To dive deeper into the problem, Figure~\ref{subfig:training loss on Molpcba} and~\ref{subfig:validation loss on Molpcba} present the training loss and validation loss curves, respectively. All the training losses in Figure~\ref{subfig:training loss on Molpcba} keep dropping in the training process. 
The larger the model is, the smaller the corresponding training loss will be. 
Nonetheless, the validation losses of models with more than $10^7$ parameters start increasing after several epochs as shown in Figure~\ref{subfig:validation loss on Molpcba}, which indicates overfitting. Moreover, the models whose performance falls behind the scaling law predictions suffer more serious overfitting. For instance, models of 7.4M, 13M, and 20M parameters overfit the most in Figure~\ref{subfig:validation loss on Molpcba}, and they are the three models on which model scaling law fails in Figure~\ref{subfig:collapse on Molpcba}.
Based on these observations, overfitting can be one potential reason for the model scaling collapse.
For more results between the model size and overfitting, please refer to Appendix~\ref{app:overfiting}.

\begin{figure}
\centering
\includegraphics[width=0.4\columnwidth]{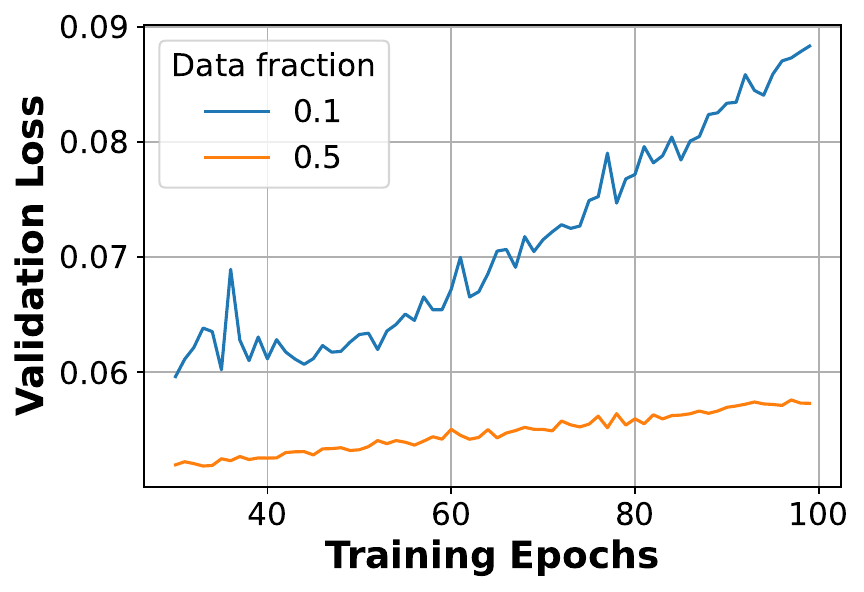}
\caption{More training data can mitigate the overfitting problem. The model size is fixed, while the two curves correspond to different data fractions of the original \molpcba{} training set.}
\label{mitigate overfitting pcba}
\end{figure}

To eliminate the effect of model scaling collapse, it is crucial to overcome the overfitting. 
\cite{kaplan2020scaling} suggests that overfitting would happen if the model size is too large for the training set. 
To investigate whether more training data can mitigate overfitting, we train a model of fixed number parameters on different fractions of the training set and illustrate the validation loss curves in Figure~\ref{mitigate overfitting pcba}. 
The model trained on $10$ percent of the original \molpcba{} dataset suffers serious overfitting as the validation loss curve suggests. 
However, the overfitting is significantly mitigated when the data fraction is raised to $50$ percent, which indicates the effectiveness of enlarging the training set. 
Another evidence is that we have not observed overfitting or model scaling collapse in models of similar sizes~($10^6$ parameters) trained on larger datasets, i.e., \PCQ{} and \ppa{} in Section~\ref{sec:Basic Laws}. 


In graph tasks, the model scaling collapse often occurs with overfitting, which is because the graph dataset sizes are much smaller compared to those in CV or NLP. 
Overall, the model scaling collapse in graph tasks suggests that the current graph datasets are small for the deep graph models, leaving the models data hungry. 
Furthermore, it indicates that a crucial step toward the large graph model is to enlarge the size of datasets.

\section{More Results on the Model Scaling Collapse\label{app:overfiting}}
Here we present more results on the relationship between the model size and the degree of overfitting. In general, the observation in the below figures indicate that the lager model is, the more serious it would overfit.

\begin{figure*}[h]
    \centering

    \includegraphics[width=0.95\columnwidth]{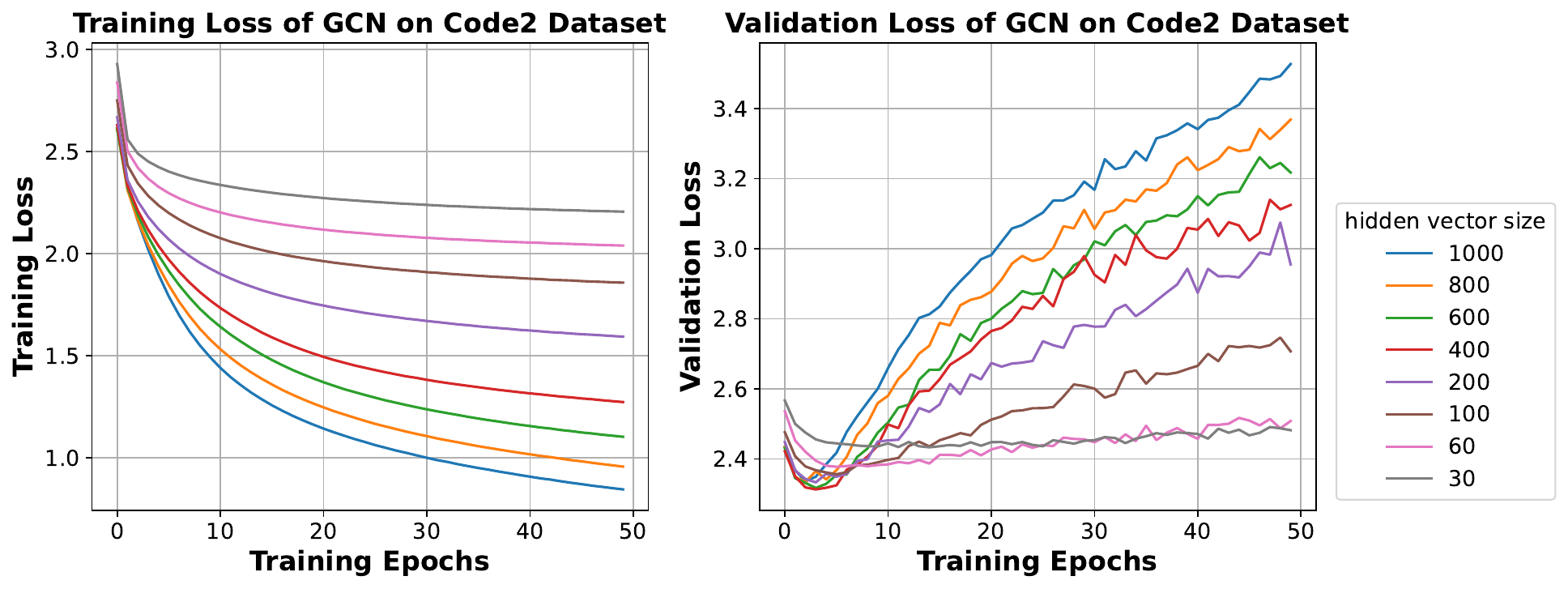}

    \caption{The training losses and validation losses of GCN with varying sizes on Code2 Dataset. The color corresponds to the number of model parameters. The figure shows that the model with larges size has more serious overfitting.}
\end{figure*}

\begin{figure*}[h]
    \centering

    \includegraphics[width=0.95\columnwidth]{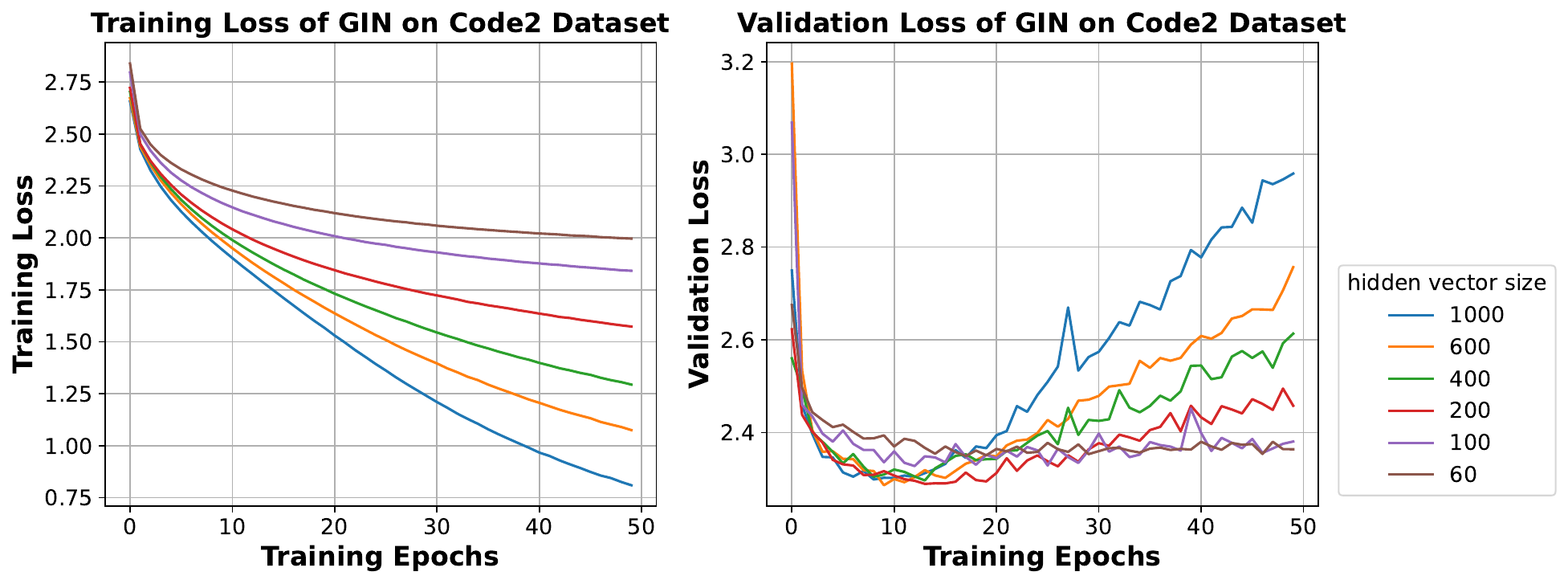}

    \caption{The training losses and validation losses of GIN with varying sizes on Code2 Dataset. The color corresponds to the number of model parameters. The figure shows that the model with larges size has more serious overfitting.}

\end{figure*}

\begin{figure*}[h]
    \centering

    \includegraphics[width=0.95\columnwidth]{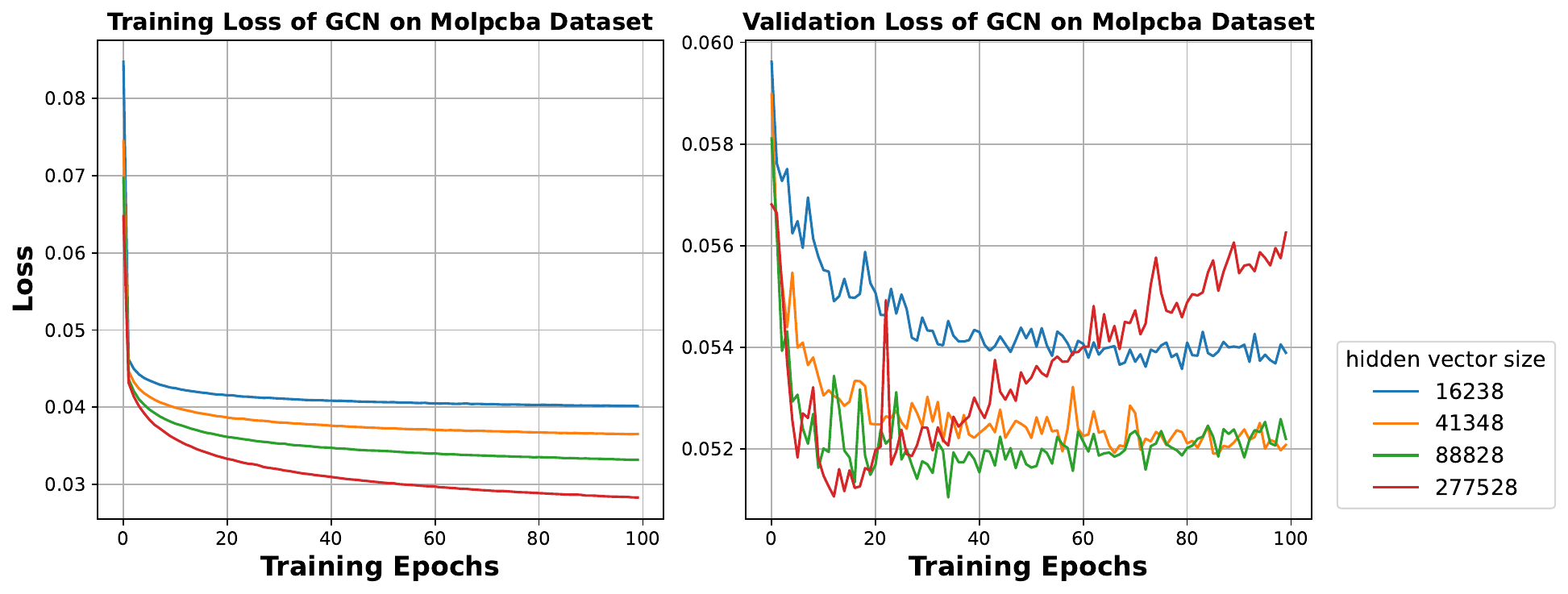}

    \caption{The training losses and validation losses of GCN with varying sizes on \ppa{} Dataset. The color corresponds to the number of model parameters. The figure shows that the model with larges size has more serious overfitting.}
\end{figure*}

\newpage
\section{More Results on the Impacts of Model Depths\label{app:model depth}}
We present the experiment results of model scaling behaviors with varying model depths on more datasets. In the following figures, models with different depths exhibit distinct model scaling behaviors.

\begin{figure*}[ht!]
    \centering
    \subfigure[Model scaling of GCN with different layers.]{
    \begin{minipage}[b]{0.45\textwidth}
    \includegraphics[width=\columnwidth]{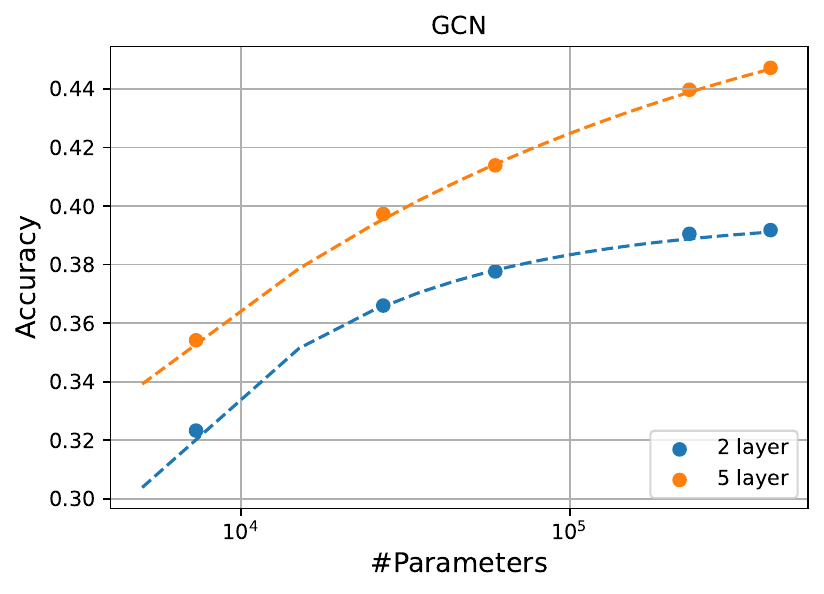}
    \end{minipage}
    }
    \subfigure[Model scaling of GIN with different layers.]{
    \begin{minipage}[b]{0.45\textwidth}
    \includegraphics[width=\columnwidth]{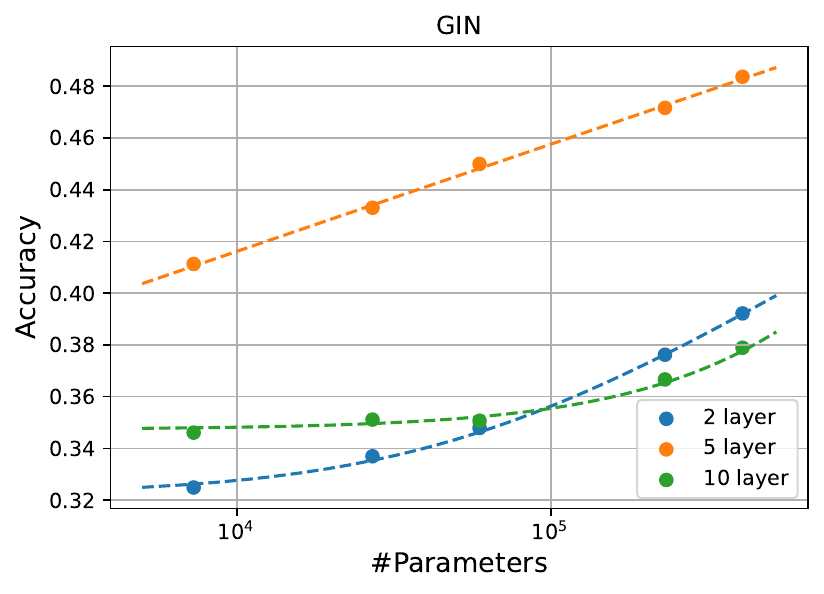}
    \end{minipage}
    }

    \caption{Model scaling behaviors of GNNs with varying model depths on \reddit{}. The color corresponds to the model layer. Both GIN and GCN exhibit distinct scaling behaviors with different model depths.}
 
\end{figure*}

\begin{figure*}[ht!]
    \centering
    \subfigure[Model scaling of GCN with different layers.]{
    \begin{minipage}[b]{0.45\textwidth}
    \includegraphics[width=\columnwidth]{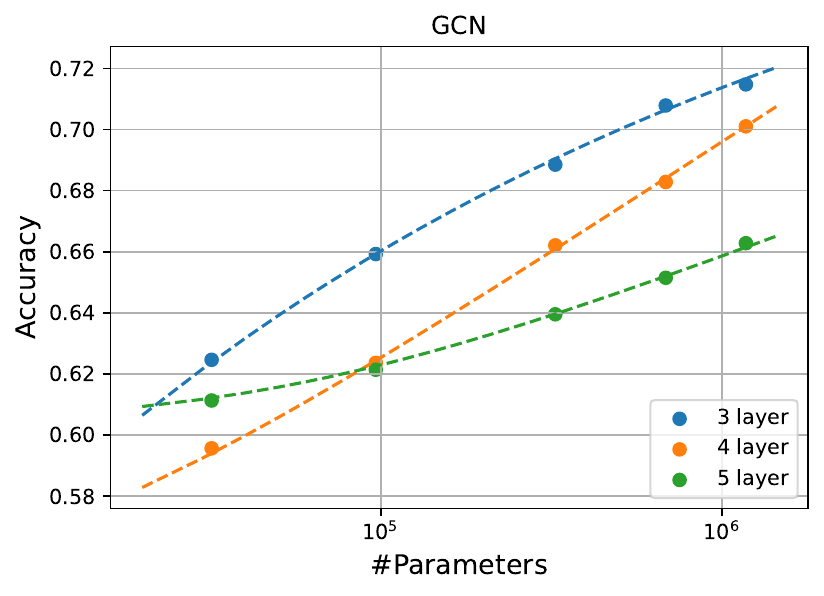}
    \end{minipage}
    }
    \subfigure[Model scaling of GIN with different layers.]{
    \begin{minipage}[b]{0.45\textwidth}
    \includegraphics[width=\columnwidth]{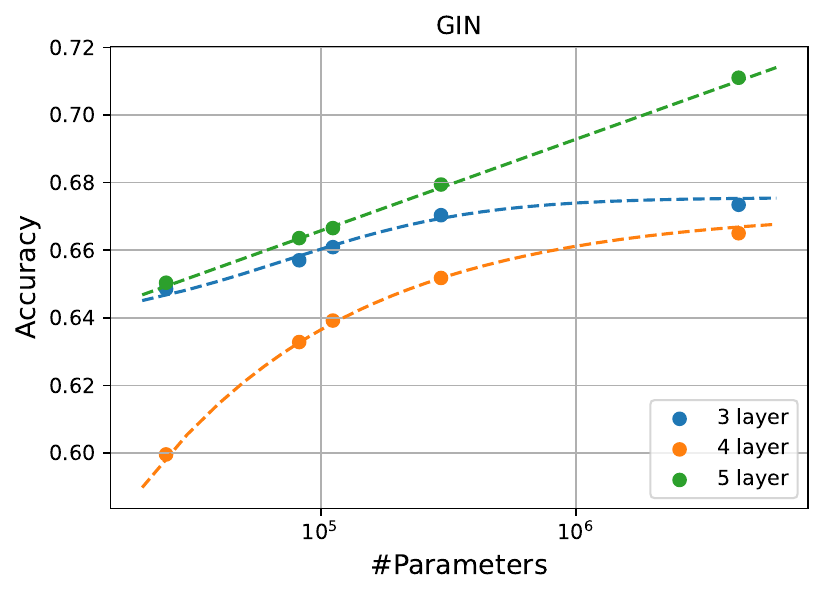}
    \end{minipage}
    }

    \caption{Model scaling behaviors of GNNs with varying model depths on \molhiv{}. The color corresponds to the model layer. Both GIN and GCN exhibit distinct scaling behaviors with different model depths.}

\end{figure*}

\begin{figure*}[ht!]
    \centering
    \subfigure[Model scaling of GCN with different layers.]{
    \begin{minipage}[b]{0.45\textwidth}
    \includegraphics[width=\columnwidth]{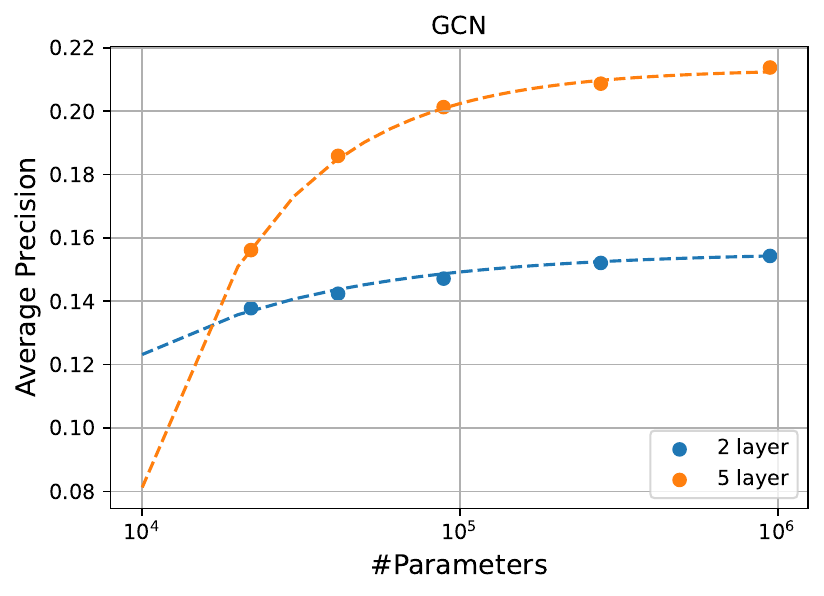}
    \end{minipage}
    }
    \subfigure[Model scaling of GIN with different layers.]{
    \begin{minipage}[b]{0.45\textwidth}
    \includegraphics[width=\columnwidth]{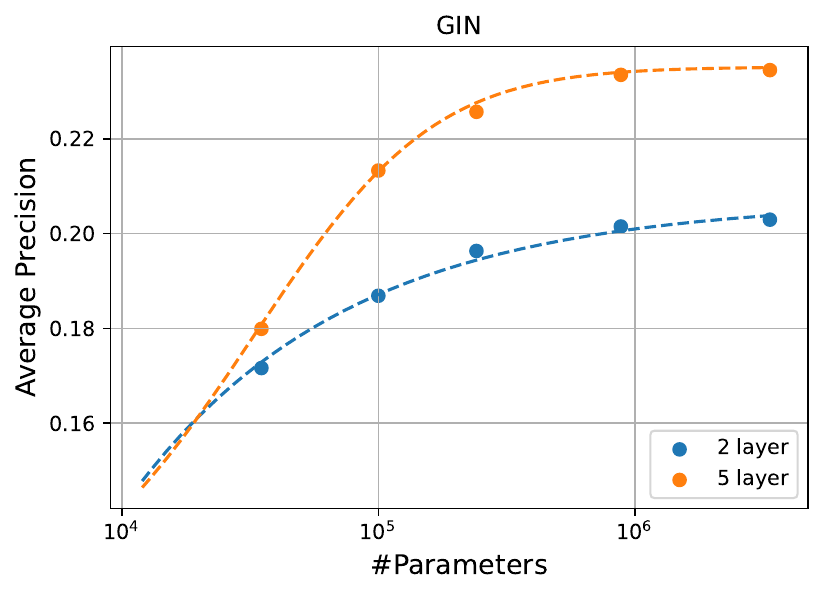}
    \end{minipage}
    }

    \caption{Model scaling behaviors of GNNs with varying model depths on \molpcba{}. The color corresponds to the model layer. Both GIN and GCN exhibit distinct scaling behaviors with different model depths.}

\end{figure*}

\vfill
\newpage
\section{The Average Degree Distributions of More Datasets\label{app:averge degree}}
\begin{figure*}[ht!]
    \centering
    \subfigure[bace]{
    \begin{minipage}[b]{0.3\textwidth}
    \includegraphics[width=\columnwidth]{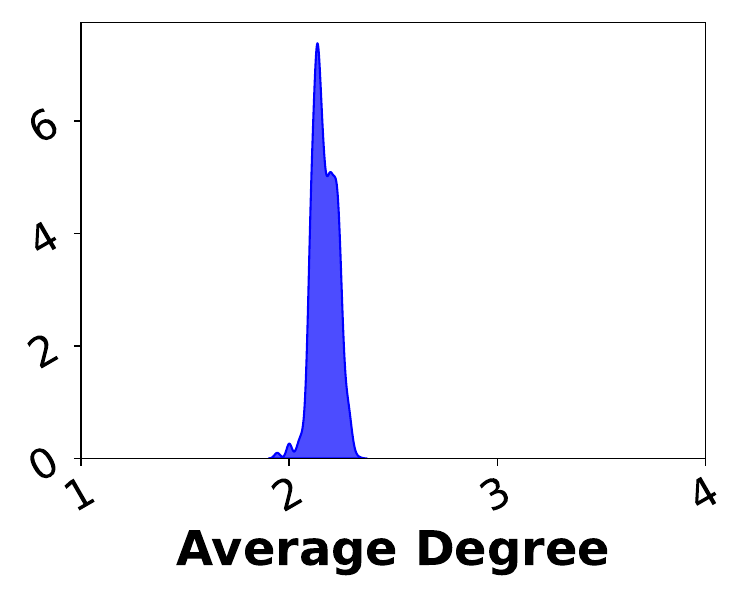}
    \end{minipage}
    }
    \subfigure[bbbp]{
    \begin{minipage}[b]{0.3\textwidth}
    \includegraphics[width=\columnwidth]{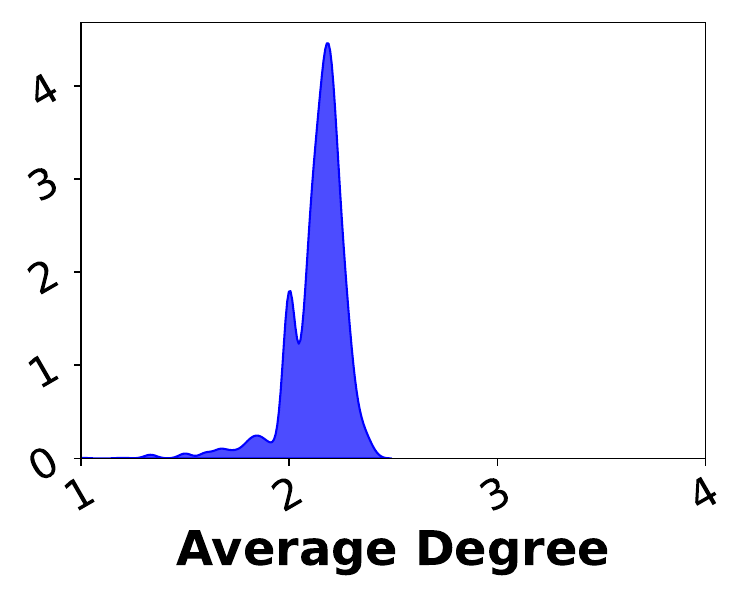}
    \end{minipage}
    }
    \subfigure[molchembl]{
    \begin{minipage}[b]{0.307\textwidth}
    \includegraphics[width=\columnwidth]{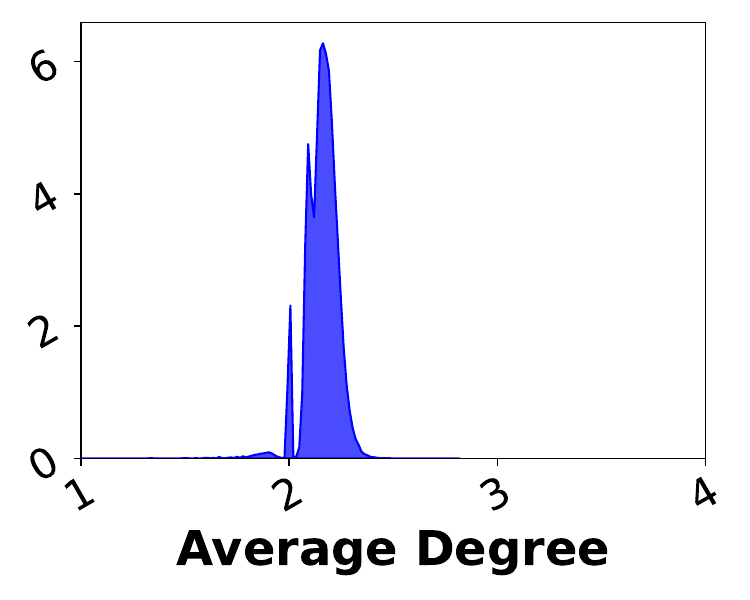}
    \end{minipage}
    }
    \vfill
    \subfigure[molpcba]{
    \begin{minipage}[b]{0.3\textwidth}
    \includegraphics[width=\columnwidth]{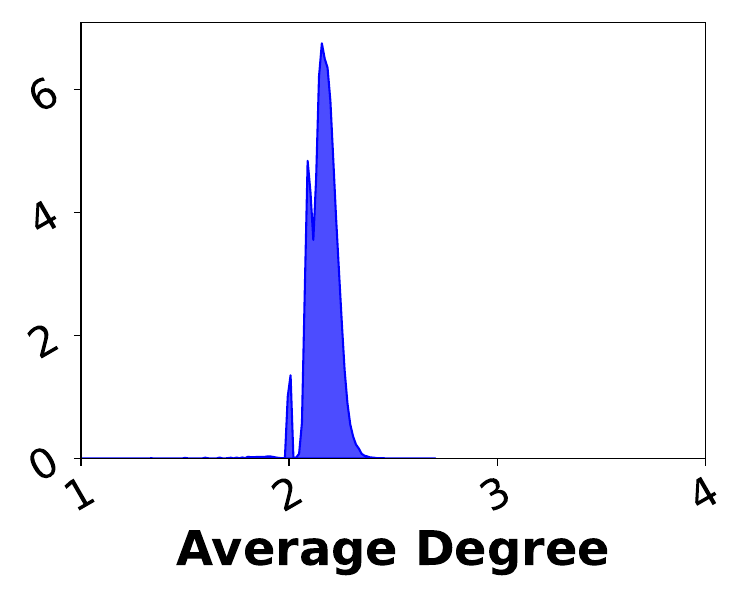}
    \end{minipage}
    }
    \subfigure[Ogbg-ppa]{
    \begin{minipage}[b]{0.3\textwidth}
    \includegraphics[width=\columnwidth]{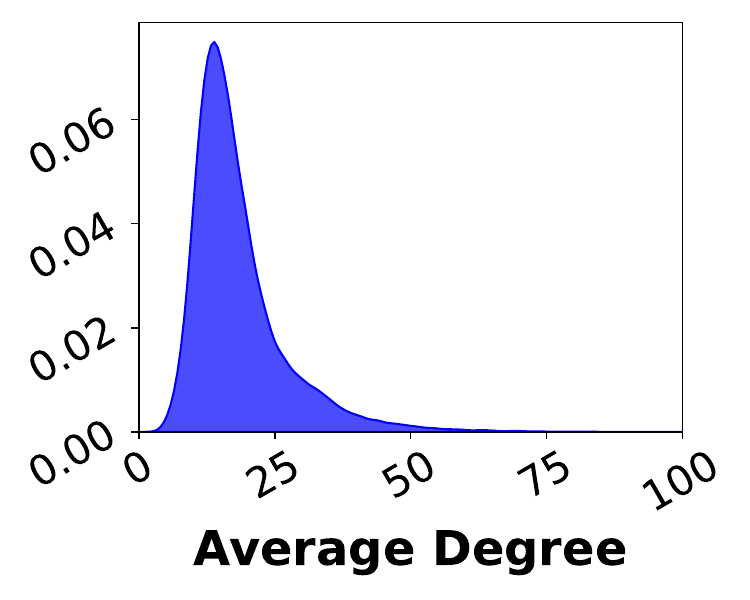}
    \end{minipage}
    }
    \subfigure[tox21]{
    \begin{minipage}[b]{0.3\textwidth}
    \includegraphics[width=\columnwidth]{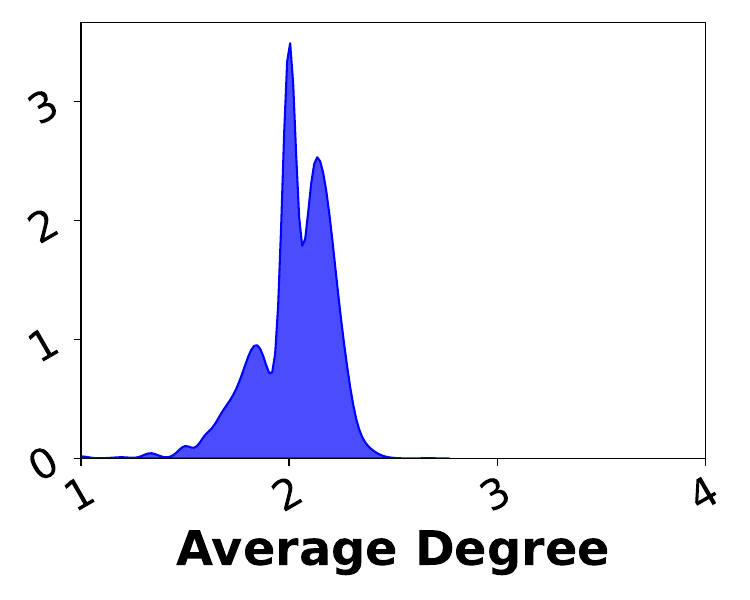 }
    \end{minipage}
    }
    \caption{The average degree distribution of more datasets.\label{fig:more average degree}}
\end{figure*}

Here we present the average degree distributions of more datasets in Figure~\ref{fig:more average degree}. The samples in one dataset tend to have similar average degrees, further supporting our argument in Section~\ref{sec:DataScaling}.

\section{Results with the Number of Nodes as metric for Section~\ref{sec:DataScaling}\label{app:results with node metric}}
\begin{figure*}[ht!]
    \centering
    \subfigure{
    \begin{minipage}[b]{0.45\textwidth}
    \includegraphics[width=\columnwidth]{fig/Synthetic_metric_graph.pdf}
    \end{minipage}
    }
    \subfigure{
    \begin{minipage}[b]{0.45\textwidth}
    \includegraphics[width=\columnwidth]{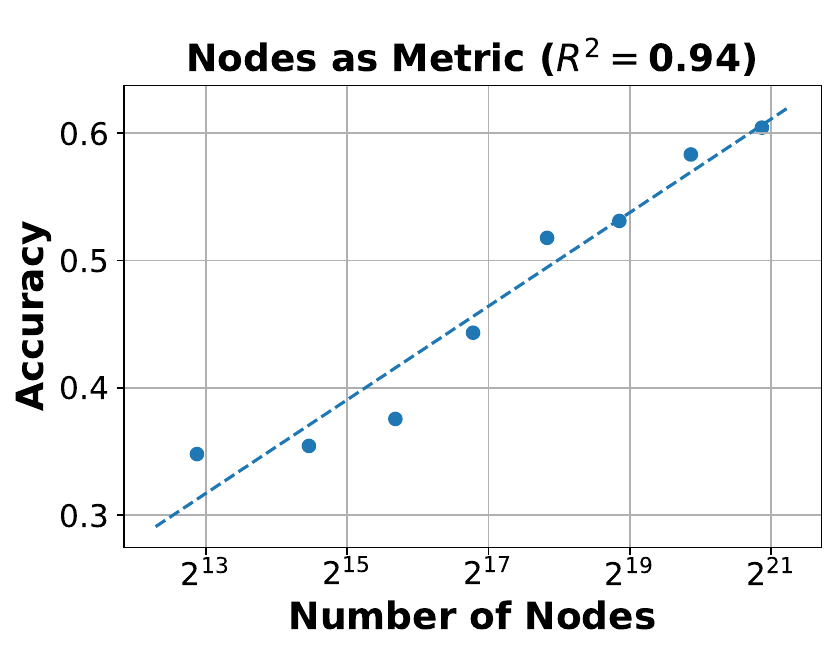}
    \end{minipage}
    }
    \caption{Comparison between the number of graphs and the number of nodes as the data metric}\label{fig:Syntheyic 2}
\end{figure*}

\begin{figure*}[ht!]
    \centering
    \subfigure{
    \begin{minipage}[b]{0.45\textwidth}
    \includegraphics[width=\columnwidth]{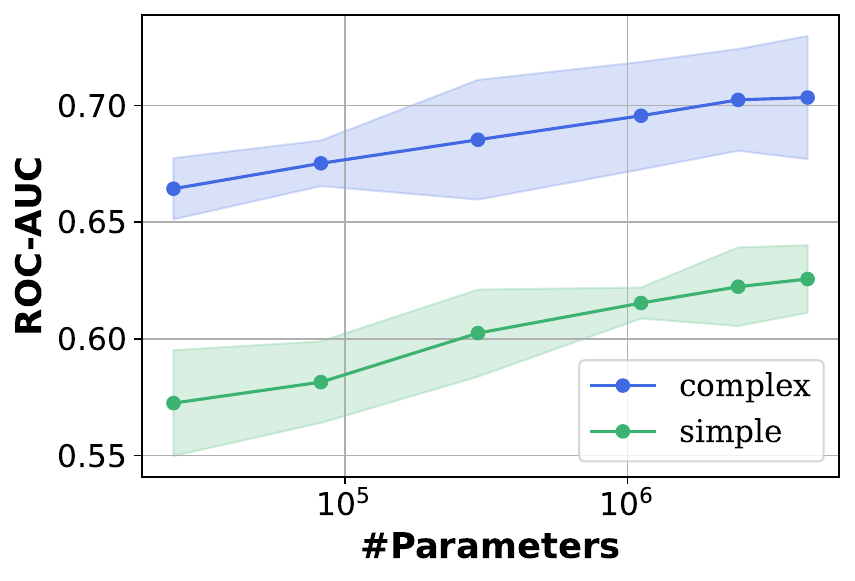}
    \end{minipage}
    }
    \subfigure{
    \begin{minipage}[b]{0.45\textwidth}
    \includegraphics[width=\columnwidth]{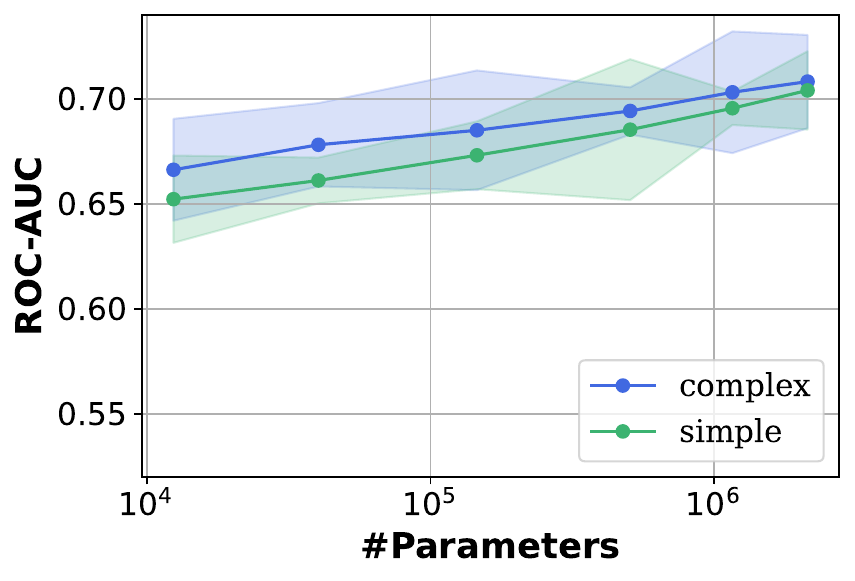}
    \end{minipage}
    }
    \caption{Model scaling curves of GIN on two training subsets of different sizes. The shade corresponds to the performance variance of repeated experiments.}\label{fig:graph number bad 2}
\end{figure*}

Here we present more results of the comparison of the data metric of the data scaling law. Figure~\ref{fig:Syntheyic 2} shows that the number of nodes as metric could also give better fitting than the number of graphs, supporting the conclusion of Experiment 1 in Section~\ref{sec:DataScaling}. Figure~\ref{fig:graph number bad 2} shows that training subsets with the same number of total nodes but different numbers of graphs tend to have similar scaling behaviors, supporting the conclusion of Experiment 2 in Section~\ref{sec:DataScaling}.

\newpage
\section{The Data Scaling Law of Node Classification and Link Prediction with the Number of Nodes as Metric\label{app:node and link with node metric}}
\begin{figure*}[ht!]
    \centering
    \subfigure{
    
    \begin{minipage}[b]{0.3\textwidth}
    \includegraphics[width=\columnwidth]{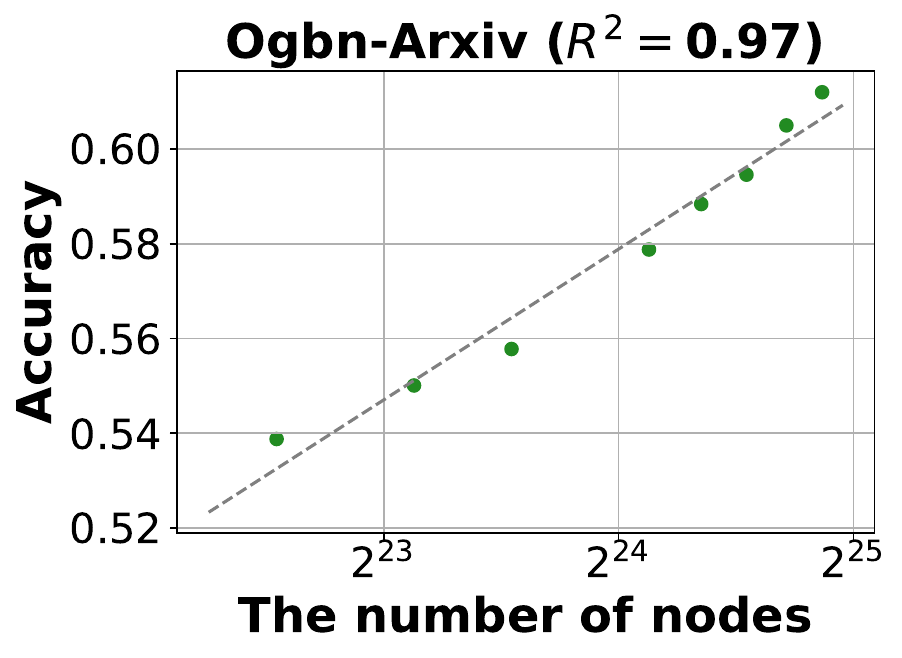}
    \end{minipage}
    }
    \subfigure{
    
    \begin{minipage}[b]{0.3\textwidth}
    \includegraphics[width=\columnwidth]{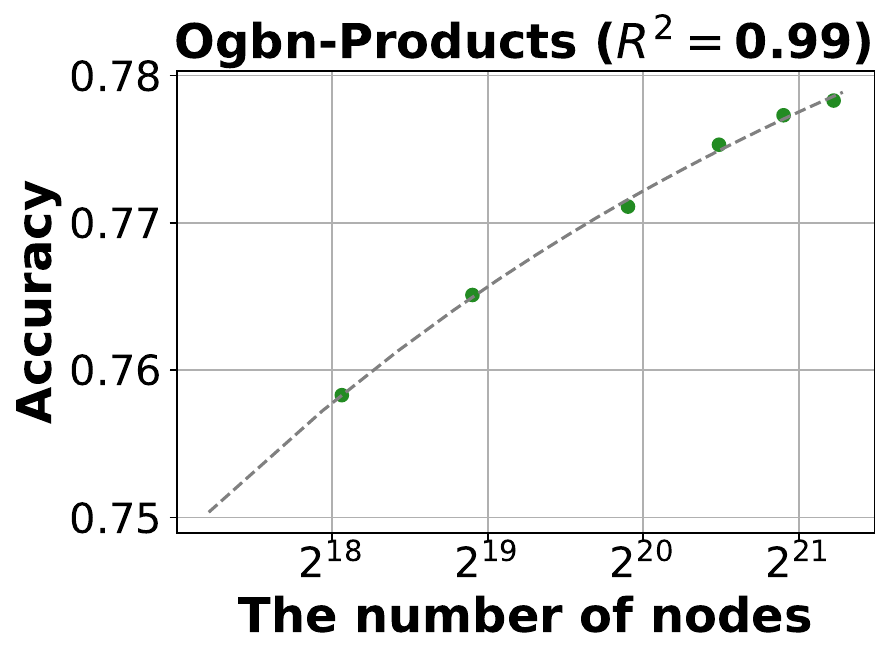}
    \end{minipage}
    }
        \subfigure{
    
    \begin{minipage}[b]{0.3\textwidth}
    \includegraphics[width=\columnwidth]{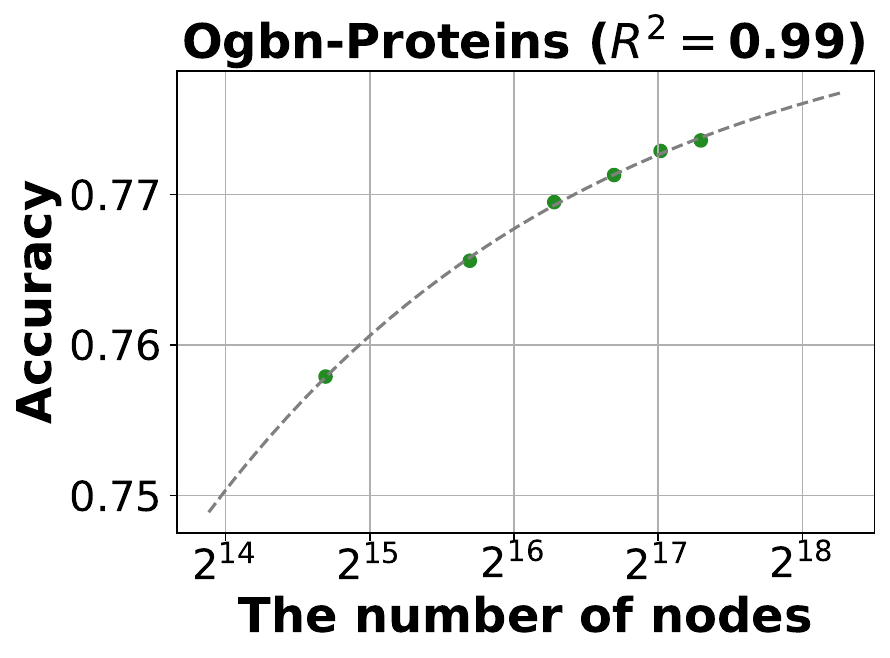}
    \end{minipage}
    }
    \vfill
    \subfigure{
    
    \begin{minipage}[b]{0.3\textwidth}
    \includegraphics[width=\columnwidth]{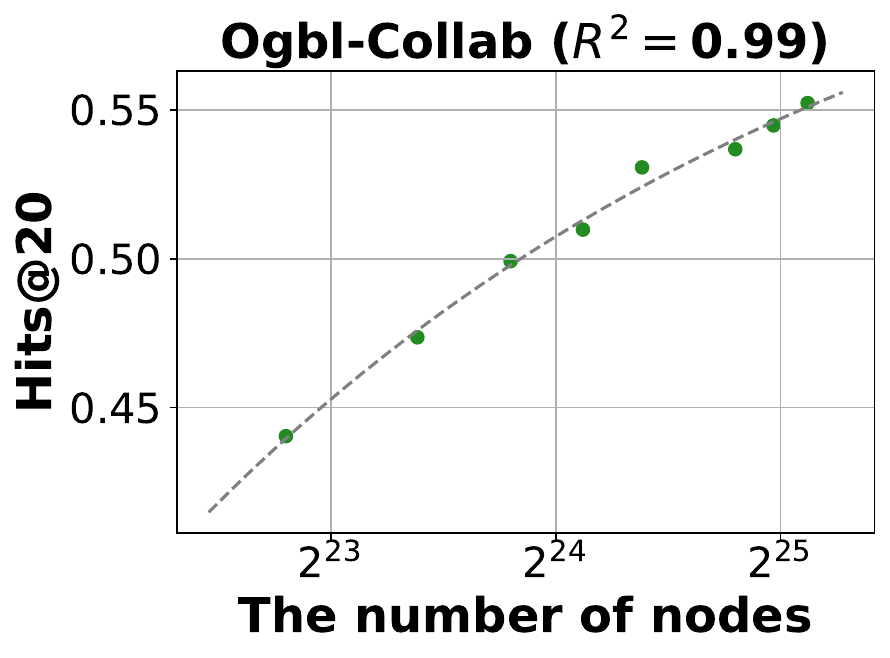}
    \end{minipage}
    }
        \subfigure{
    
    \begin{minipage}[b]{0.3\textwidth}
    \includegraphics[width=\columnwidth]{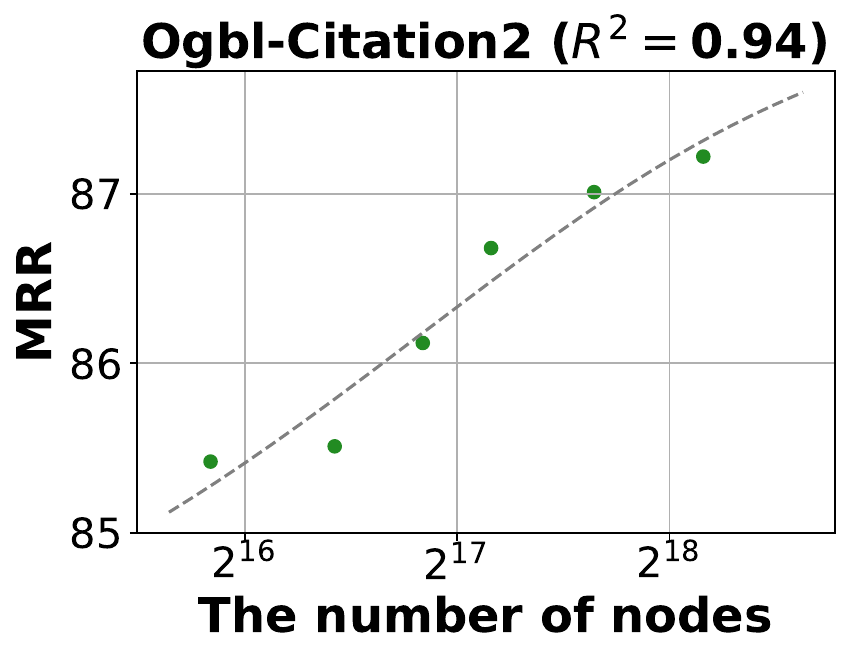}
    \end{minipage}
    }
            \subfigure{
    
    \begin{minipage}[b]{0.3\textwidth}
    \includegraphics[width=\columnwidth]{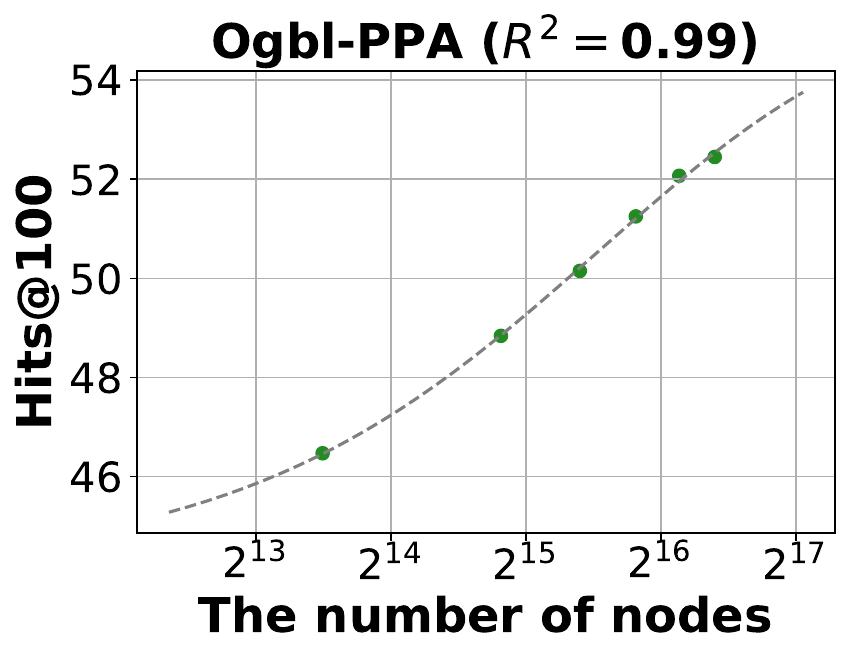}
    \end{minipage}
    }

    \caption{Both node classification and link prediction follow the data scaling law with the total node number as the data metric.}\label{fig:node and link with node metric}

\end{figure*}

Here we present the results of the data scaling law on node classification and link prediction tasks with the number of nodes as the data metric. From Figure~\ref{fig:node and link with node metric}, we find that the number of nodes also gives good fitting results as the data metric.
\newpage
\section{How do Our Findings Connect to Compute Scaling and Chinchilla Scaling Law\label{app:edge and flops}}

Here we first show that the training compute can be calculated when we know the total edge number and the model architecture. We consider the general Message Passing Neural Networks, which can be formulated as above:
\begin{equation}
    m^{t+1}_v = U(h^t_v, \sum_{w\in N(v)} M(h^t_v, h^t_w, e_{vw}))
\end{equation}
where $U$ and $M$ can be assumed to contain $x$ and $y$ add-multiply operations, respectively. Suppose the total node number in the dataset is N, the total edge number is E, and the model has L layers.
For node $i$, aggregating messages from its neighbors needs $x+yd_i$ add-multiply operations, where $d_i$ is the degree of node $i$. For the whole datasets, there would be $\sum_i x+yd_i= 2(1+1/d_a)yE$, where $d_a$ is the average degree of the dataset. For the whole model, one feed-forward pass needs $2(1+1/d_a)yEL$ add-multiply operations. Thus, we have shown that we can calculate the training FLOPs if we know the total edge number in the dataset.

Chinchilla scaling law aims to allocate the best training set size and the model size when the compute cost is fixed~\cite{hoffmann2022training}.
The graph scaling law developed in our paper can be easily extended to calculate optimal parameter numbers or dataset size as shown below. For simplicity, we just consider Equation 4 here. But the results can be easily extended to Equation 5. For Equation 4 we have $$\epsilon = \epsilon_\infty + A/N^\alpha + B/D^\beta$$ where N is the model size, and D is the training set size. From the discussion above and in~\cite{kaplan2020scaling}, we know the FLOPs for GIN are about $6\frac{1+d_a}{d_a}EN$, where E is the number of edges and $d_a$ is the average degree of the training set. For fixing the compute budget $C = 6\frac{1+d_a}{d_a}EN$, we can get $N_{opt} = G(\frac{d_aC}{6(1+d_a)})^a$, $E_{opt} = G^{-1}(\frac{d_aC}{6(1+d_a)})^b$ following [2], where $G = (\frac{\alpha A}{\beta B})^{\frac{1}{\alpha+\beta}}$, $a = \frac{\beta}{\alpha+\beta}$, $b = \frac{\alpha}{\alpha+\beta}$. We can find that there is a power-law relationship between the optimal parameters/ training set size and the FLOPs.

\section{The Model Scaling Behaviors of Node Classification and Link Prediction\label{app:nodel and link}}
Here we investigate the model scaling behaviors of node classification and link prediction tasks. The observations in the below figure show that SEAL trained on \collab{} has obvious performance gains when the model size is increased, while this is not the case for GraphSAGE on \arxiv{}.
We leave further exploration of these topics for future works.

\begin{figure*}[ht!]
    \centering
    \subfigure[Model scaling behaviors on \arxiv{}.]{
    \begin{minipage}[b]{0.4\textwidth}
    \includegraphics[width=\columnwidth]{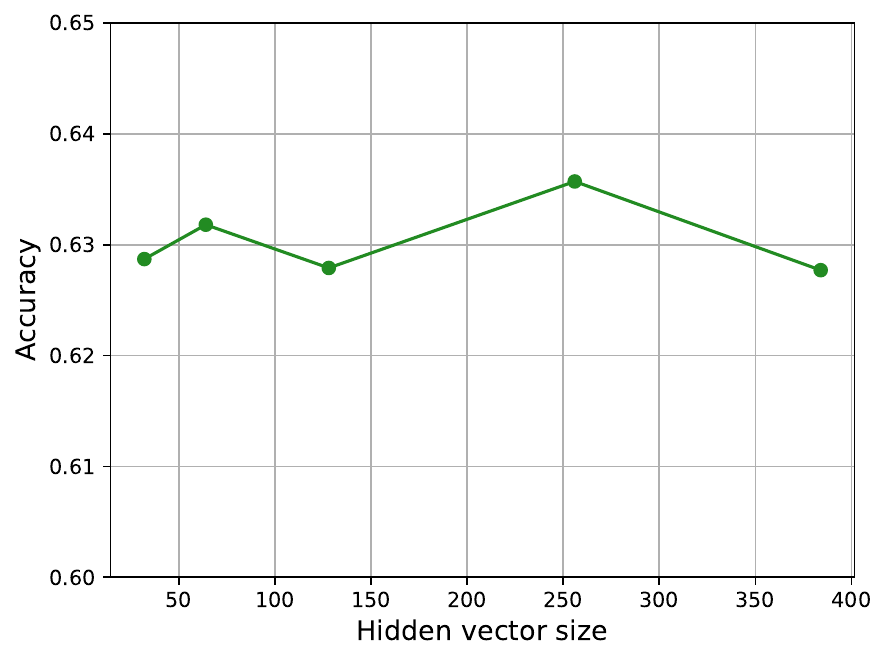}
    \end{minipage}
    }
    \subfigure[Model scaling on \collab{}.]{
    \begin{minipage}[b]{0.4\textwidth}
    \includegraphics[width=\columnwidth]{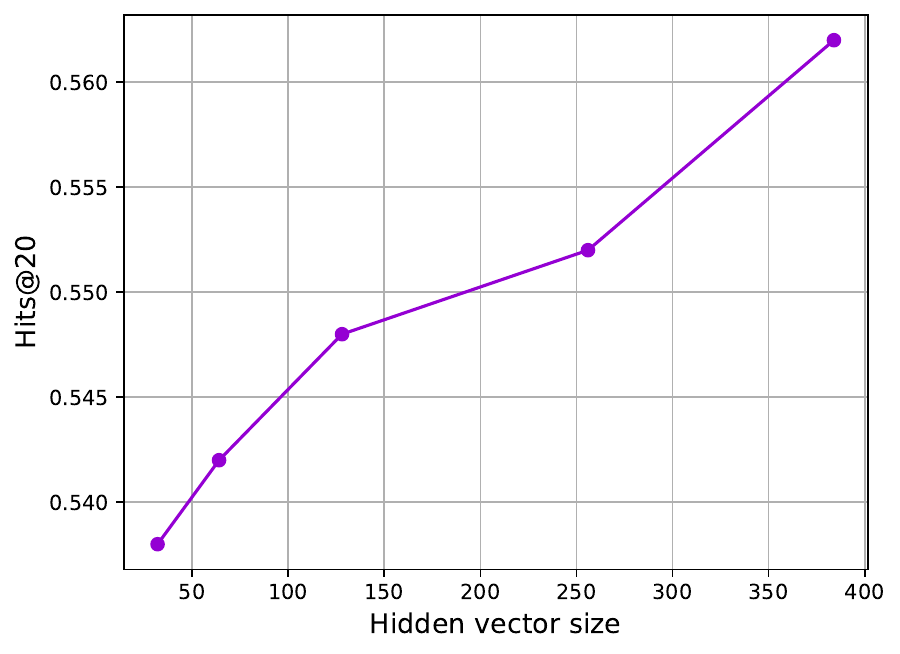}
    \end{minipage}
    }

    \caption{In (b), the model performance has obvious gains on the link prediction task when the hidden vector size is increased, while it is not for the node classification task in (a).}

\end{figure*}
\section{Broader Impacts}
In this paper, we investigate neural scaling laws on graphs, which describe how the model performance will grow with increasing model and data sizes.
Neural scaling laws can provide instructions for building large graph models, predicting the costs of computing and data labeling to achieve the expected performance.
Moreover, it may incentivize more investments in data collecting and even the construction of larger graph datasets for various tasks.
We anticipate that our findings will serve as essential guidance for the development of large graph models.

\end{document}